\documentclass{article}

\usepackage[utf8]{inputenc} % allow utf-8 input
\usepackage[T1]{fontenc}    % use 8-bit T1 fonts
\usepackage[colorlinks=true, citecolor=blue, urlcolor=magenta]{hyperref}
\usepackage{url}            % simple URL typesetting
\usepackage{booktabs}       % professional-quality tables
\usepackage{amsfonts}       % blackboard math symbols
\usepackage{nicefrac}       % compact symbols for 1/2, etc.
\usepackage{microtype}      % microtypography
\usepackage{xcolor}         % colors
\usepackage{makecell}
\usepackage{diagbox}
\usepackage{graphicx}
\usepackage{wrapfig} 
\usepackage{wrapfig} 
\usepackage{caption}
\usepackage{subcaption}
\PassOptionsToPackage{numbers}{natbib}
\usepackage{amsmath}
\usepackage{amssymb}
\usepackage{textcomp}
\usepackage{multirow}
\usepackage{listings}
\usepackage[dandb]{neurips_2025}
\usepackage[most]{tcolorbox}
\definecolor{myGreen1}{RGB}{210, 238, 255}   % 浅蔚蓝 (AliceBlue)
\definecolor{myYellow}{RGB}{ 251, 209,188}  
\definecolor{myYellow2}{RGB}{ 245, 141,88}   
\definecolor{darkblue}{RGB}{17, 136, 228} % 深蓝色（navy blue）

\newtcolorbox{RoundedBox}[2][]{%
  colback=myGreen1,              % 内容区浅蓝
  colframe=blue!65!black,      % 边框深蓝
  boxrule=0.8pt,               % 边框粗细
  arc=4pt, outer arc=4pt,      % 圆角尺寸
  coltitle=white,              % 标题文字白色
  colbacktitle=blue!65!black,  % 标题背景同边框色
  fonttitle=\bfseries\small,   % 标题字体加粗并设小号
  title=#2,                    % 标题内容
  left=4pt, right=4pt,
  top=4pt, bottom=4pt,
  #1
}

\title{STEP: A Unified Spiking Transformer Evaluation Platform for Fair and Reproducible Benchmarking}

\author{
  Sicheng Shen$^{1,2,4,\bigstar}$ \quad
  Dongcheng Zhao$^{1,3,\bigstar}$ \quad
  Linghao Feng $^{1}$ \\
  \textbf{Zeyang Yue}$^{1,5}$ \quad
  \textbf{Jindong Li}$^{1}$ \quad 
  \textbf{Tenglong Li}$^{1}$ \\
  \textbf{Guobin Shen}$^{1}$\quad
  \textbf{Yi Zeng}$^{1,3,\dag}$\\
   $^1$ BrainCog Lab, CASIA \quad
  $^2$ School of Future Tech., UCAS \quad
  $^3$ Long-term AI \\
  $^4$ Zhongguancun Academy \quad
  $^5$ Beihang University \\
  $^\bigstar$ Equal contribution \quad
  $^\dag$ Corresponding author \quad
  \\
  \href{mailto:shensicheng2024@ia.ac.cn}{\textcolor{darkblue}{\texttt{shensicheng2024@ia.ac.cn}}} \quad
  \href{mailto:yi.zeng@ia.ac.cn}{\textcolor{darkblue}{\texttt{yi.zeng@ia.ac.cn}}}
}

\begin{document}

\maketitle

\begin{abstract}
Spiking Transformers have recently emerged as promising architectures for combining the efficiency of spiking neural networks with the representational power of self-attention. However, the lack of standardized implementations, evaluation pipelines, and consistent design choices has hindered fair comparison and principled analysis. In this paper, we introduce \textbf{STEP}, a unified benchmark framework for Spiking Transformers that supports a wide range of tasks, including classification, segmentation, and detection across static, event-based, and sequential datasets. STEP provides modular support for diverse components such as spiking neurons, input encodings, surrogate gradients, and multiple backends (e.g., SpikingJelly, BrainCog). Using STEP, we reproduce and evaluate several representative models, and conduct systematic ablation studies on attention design, neuron types, encoding schemes, and temporal modeling capabilities. We also propose a unified analytical model for energy estimation, accounting for spike sparsity, bitwidth, and memory access, and show that quantized ANNs may offer comparable or better energy efficiency. Our results suggest that current Spiking Transformers rely heavily on convolutional frontends and lack strong temporal modeling, underscoring the need for spike-native architectural innovations.  \textbf{The full code is available at:} \url{https://github.com/Fancyssc/STEP}.
\end{abstract}

\section{Introduction}

Spiking Neural Networks (SNNs) are a biologically inspired paradigm that simulate neural information processing via discrete spikes. These networks excel not only at static image tasks but also in modeling dynamic and temporally structured data~\cite{maass1997networks}. Their event-driven nature contributes to high energy efficiency and strong biological plausibility. However, applying SNNs to deep learning architectures—particularly Transformers—remains challenging due to their non-differentiability, limited scalability, and training instability.

In parallel, Artificial Neural Networks (ANNs) have seen tremendous advances through architectural innovations. ResNet~\cite{he2016deep} introduced residual learning to ease optimization in deep networks, while Recurrent Neural Networks (RNNs) captured sequential dependencies. The Transformer architecture~\cite{vaswani2017attention} unified these advances by leveraging self-attention, enabling efficient parallel modeling of long-range dependencies. Vision Transformer (ViT)~\cite{dosovitskiy2020image} further demonstrated the potential of attention mechanisms in visual tasks. Drawing inspiration from these architectures, the SNN community has proposed Spiking ResNet~\cite{hu2021spiking}, SEW-ResNet~\cite{fang2021deep}, and spiking RNN variants~\cite{xing2020new,xu2024rsnn}. Recently, attention-based spiking models such as Spikformer~\cite{zhou2022spikformer}, QKFormer~\cite{zhou2024qkformer}, and SpikingResformer~\cite{shi2024spikingresformer} have emerged. The Spike-Driven Transformer series~\cite{yao2023spike,yao2024spike,yao2025scaling} improves both efficiency and scalability, enabling applications in image segmentation and object detection.

% Despite these advancements, several key challenges persist. First, the performance gap between spiking Transformers and traditional ANNs remains unclear, particularly regarding their unique advantages on temporal or event-based data. Systematic evaluations across static (e.g., ImageNet), event-driven (e.g., DVS-CIFAR10), and sequential (e.g., SCIFAR10) datasets are needed to rigorously assess their potential. Second, Spiking Transformers consist of multiple interacting components—such as spike encoders, neuron models, surrogate gradients, attention modules, and MLP heads—yet the contribution of each module remains underexplored. Module-wise ablation is crucial to understanding design trade-offs and guiding model optimization. Third, while SNNs offer inherent energy benefits through sparse, binary spike-based computation, direct comparisons to quantized Transformers are rare. Quantifying the energy-performance trade-off is essential for evaluating the practical utility of spiking models.

% Moreover, inconsistencies across development frameworks further hinder progress. Existing tools such as SpikingJelly~\cite{fang2023spikingjelly}, BrainCog~\cite{zeng2023braincog}, and BrainPy~\cite{wang2023brainpy} differ in neuron implementations, training protocols, and evaluation interfaces. These inconsistencies complicate reproducibility, hyperparameter tuning, and fair model comparison. Currently, there is no unified platform for evaluating Spiking Transformers across tasks such as classification, segmentation, and detection.

% 精简
Despite advancements, several key challenges persist in Spiking Transformers (STs). First, the performance gap between STs and traditional ANNs remains unclear, especially regarding their unique advantages on temporal or relatively complicated data. A systematic evaluation across diverse datasets—static (e.g., ImageNet), event-driven (e.g., DVS-CIFAR10), and sequential (e.g., SCIFAR10)—is essential for assessing their potential. Second, STs consist of multiple interacting components, including spike encoders, neuron models, surrogate gradients, attention modules, and MLP heads, yet the contribution of each module is underexplored. Module-wise ablation is critical for understanding trade-offs and guiding optimization. Third, while SNNs inherently offer energy benefits through sparse, binary spike-based computation, direct comparisons to quantized Transformers are scarce. Quantifying the energy-performance trade-off is necessary to assess the practical utility of spiking models. Moreover, inconsistencies across development frameworks, such as SpikingJelly~\cite{fang2023spikingjelly}, BrainCog~\cite{zeng2023braincog}, and BrainPy~\cite{wang2023brainpy}, further hinder progress by complicating reproducibility, hyperparameter tuning, and fair model comparison. Currently, no unified platform exists for evaluating Spiking Transformers across tasks like classification, segmentation, and detection.

To address these challenges, we introduce the \textbf{Spiking Transformer Evaluation Platform (STEP)}, a unified benchmarking framework for building, evaluating, and comparing Spiking Transformers. STEP integrates representative implementations, supports modular component replacement, and enables consistent evaluation across visual tasks. It provides both training-from-scratch and pretraining–finetuning pipelines, and supports integration with backends such as SpikingJelly, BrainCog, and BrainPy. Moreover, leveraging MMSegmentation~\cite{mmseg2020} and MMDetection~\cite{mmdetection}, STEP extends support to dense prediction tasks. Our main contributions are as follows:
\begin{itemize}
    \item We propose a unified benchmarking framework (STEP) for Spiking Transformers, integrating existing implementations to ensure consistency and reproducibility in evaluation.
    \item We design module-wise ablation experiments to evaluate the contribution of core components, providing guidance for architectural optimization.
    \item We investigate energy–performance trade-offs between Spiking and quantized Transformers, highlighting the unique advantages of spike-based computation.
\end{itemize}

 \section{Preliminary}

\begin{figure}[t]
\centering
\includegraphics[width=0.9\textwidth]{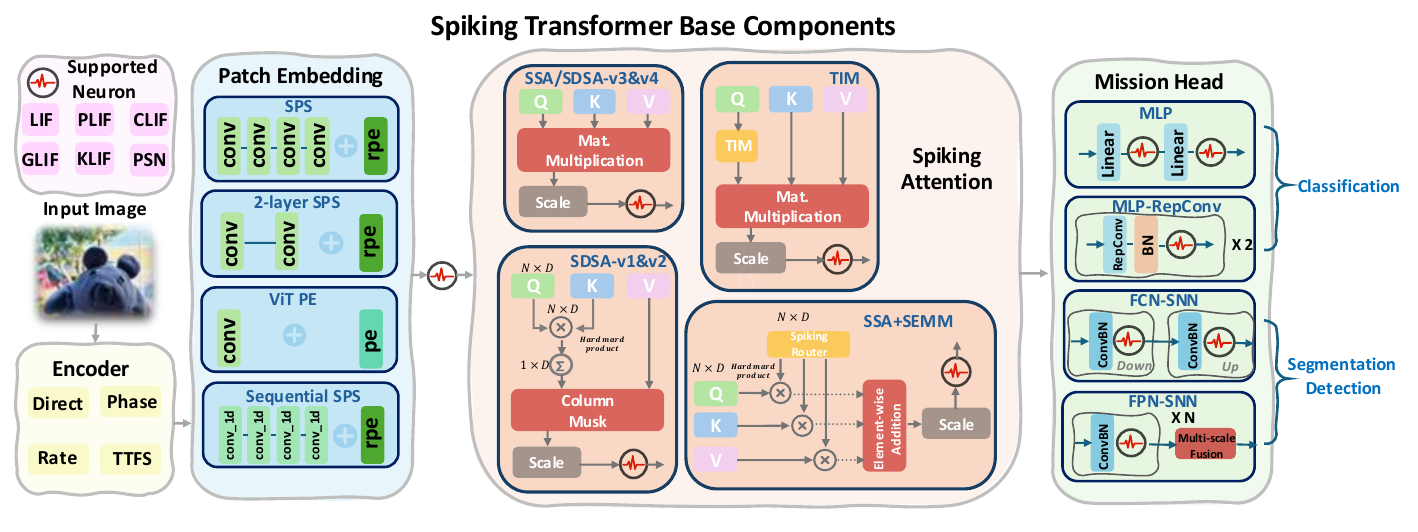}
\caption{Unified Spiking Transformer Framework with Flexible Encoding, Attention Modules, and Application-specific Heads}
\label{fig: st component}
\end{figure}

Spiking Transformers (STs) integrate the sparse, event-driven processing of Spiking Neural Networks (SNNs) with the scalable representation power of Transformer architectures (Fig.~\ref{fig: st component}). This hybrid design enables efficient handling of static and dynamic data, benefiting from both energy efficiency and long-range contextual modeling. Key components of STs include spike-based input encoding, spiking neurons, patch-wise tokenization, position embeddings, spiking self-attention (SSA), and task-specific prediction heads.

\paragraph{SNN Input Encoding}
To enable spike-based processing, input signals are transformed into temporal spike trains via encoding schemes such as direct, rate, time-to-first-spike (TTFS), and phase encoding~\cite{adrian1926impulses, park2020t2fsnn, kim2018deep}. A detailed overview of encoding methods is provided in Appendix~\ref{ap:encoding}.
\paragraph{Spiking Neurons}
Spiking neurons transmit information via discrete spikes triggered by membrane potential dynamics. The Leaky Integrate-and-Fire (LIF) model~\cite{hunsberger2015spiking} is widely used due to its simplicity and biological plausibility:
\begin{equation}
V[t] = V[t-1] + \frac{1}{\tau} (X[t] - V[t-1]), \quad \text{if } V[t] \geq V_{th}, \text{ emit spike and reset.}
\end{equation}
Variants like PLIF~\cite{fang2021incorporating} and GLIF~\cite{yao2022glif} enhance adaptability with learnable decay or gated mechanisms. Further details are provided in Appendix~\ref{ap:neuron}.

\paragraph{Spiking Self-Attention}
SSA adapts the attention mechanism to the spike domain, enabling long-range dependencies without softmax. Given input $X$, SSA computes spiking queries, keys, and values:
\begin{equation}
Q = SN_Q(W_Q^\top X), \quad K = SN_K(W_K^\top X), \quad V = SN_V(W_V^\top X), \quad SSA = SN(QK^\top V) \cdot \text{scale}
\end{equation}
Here, $SN(\cdot)$ denotes selected spiking neuron. This mechanism preserves temporal sparsity while capturing global context. See Appendix~\ref{ap:Attn} for SSA variants.

\paragraph{Other Modules}
Patch-based tokenization (Spiking Patch Splitting) enables scalable input decomposition, while Position Embeddings inject spatial/temporal order into spike sequences. Final predictions are made via MLP heads adapted for classification, detection, or segmentation.

\paragraph{Recent Advancements}
Recent ST models propose lightweight attention~\cite{yao2023spike, yao2024spike}, hierarchical designs (e.g., QKFormer~\cite{zhou2024qkformer}), and multi-task heads (e.g., FCN~\cite{long2015fully}, FPN~\cite{lin2017feature}) to enhance performance across modalities. These improvements drive STs toward practical deployment while retaining neuromorphic efficiency.

\section{Spiking Transformer Benchmark}

Building on the core components of Spiking Transformers, we present the \textbf{Spiking Transformer Evaluation Platform (STEP)}—a unified, extensible benchmark designed to standardize evaluation and accelerate research in this emerging field. STEP supports a wide range of tasks, including classification, segmentation, and object detection, and enables fair, reproducible comparisons across different models and datasets.

\begin{figure}[h]
\centering
\includegraphics[width=0.75\textwidth]{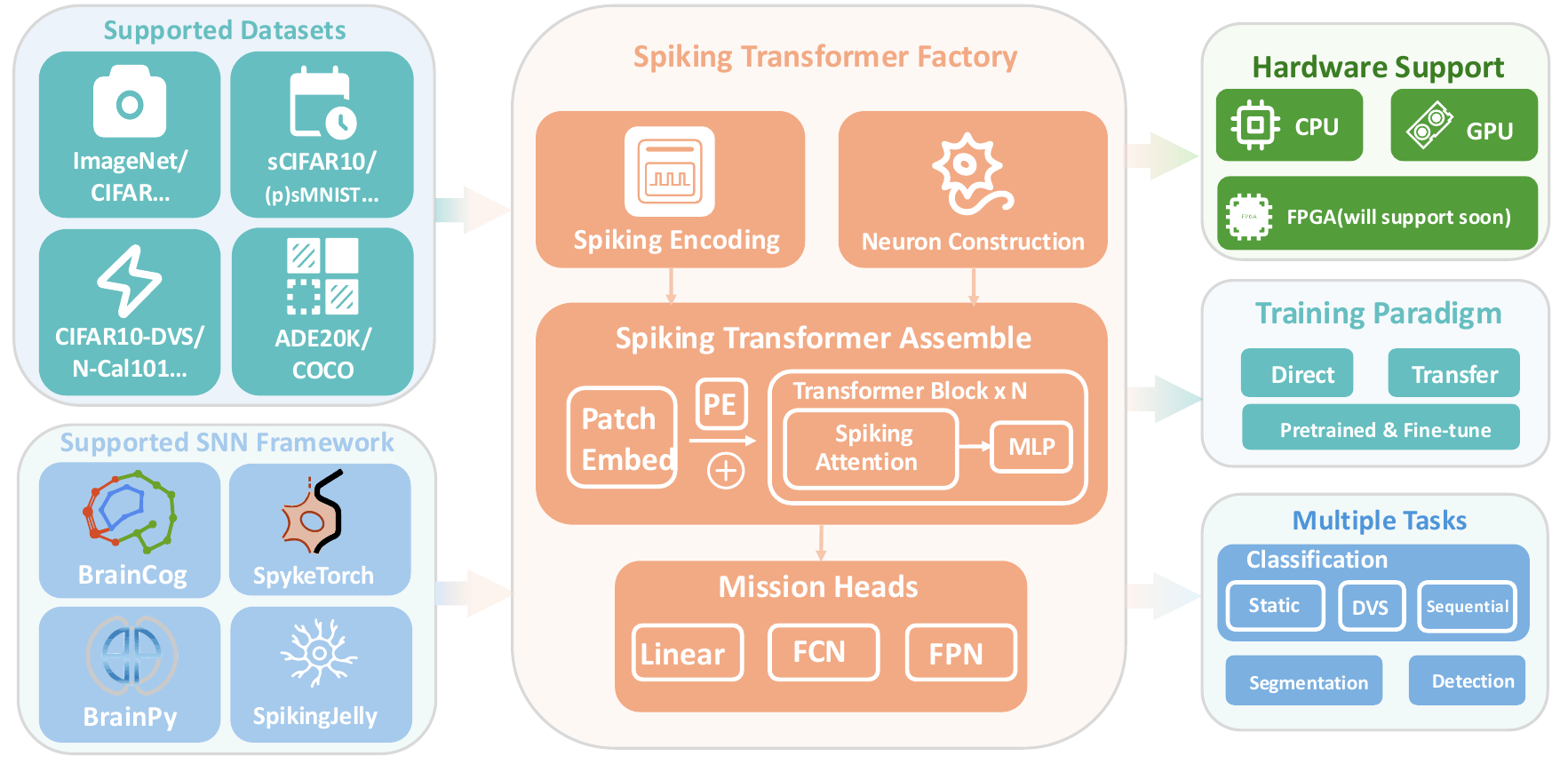}
\caption{System Architecture of STEP as a Unified Benchmark for Spiking Transformer Development and Evaluation}
\label{fig: step}
\end{figure}

STEP is built around four key principles (Fig.~\ref{fig: step}): (1) \textit{modularity}, allowing flexible integration of neuron models, encodings, and attention mechanisms; (2) \textit{dataset compatibility}, supporting static, event-based, and sequential inputs; (3) \textit{multi-task adaptation}, with pipelines for vision tasks beyond classification; and (4) \textit{backend interoperability}, enabling seamless deployment across major SNN frameworks such as SpikingJelly, BrainCog, and BrainPy.

Together, these design goals make STEP a robust foundation for developing, benchmarking, and extending Spiking Transformers. It not only reduces implementation overhead but also helps identify architectural bottlenecks and promotes best practices, fostering progress toward more generalizable and practical neuromorphic models. For detailed usage instructions, please refer to the Appendix~\ref{ap:step_quickstart}.

% Building on Spiking Transformers, we present the \textbf{Spiking Transformer Evaluation Platform (STEP)}, a unified, extensible benchmark to standardize evaluation and accelerate research in this field. STEP supports classification, segmentation, and detection tasks, enabling fair, reproducible comparisons across models and datasets.

% STEP is designed around four principles (Fig.~\ref{fig: step}): (1) \textit{modularity}, enabling flexible integration of neuron models, encodings, and attention mechanisms; (2) \textit{dataset compatibility}, supporting static, event-based, and sequential inputs; (3) \textit{multi-task adaptation}, with pipelines for vision tasks beyond classification; (4) \textit{backend interoperability}, ensuring seamless deployment across SNN frameworks like SpikingJelly, BrainCog, and BrainPy.

% These goals make STEP a solid foundation for developing, benchmarking, and extending Spiking Transformers, reducing implementation overhead, identifying architectural bottlenecks, and promoting best practices for more generalizable, practical neuromorphic models.

\subsection{Flexible and Modular Architecture}

The Spiking Transformer Benchmark is designed with a modular and extensible architecture that supports seamless integration across various backend frameworks. It accommodates diverse neuron models, encoding schemes, and surrogate gradients, allowing researchers to tailor the benchmark to specific design requirements or research goals. A unified training pipeline ensures consistent evaluation protocols, while the low-coupling structure enables independent modification of core components such as patch embedding, attention mechanisms, and MLP heads.

\subsection{Broad Dataset Compatibility}

Our benchmark supports a wide spectrum of datasets, encompassing static (e.g., ImageNet~\cite{deng2009imagenet}), event-based (e.g., DVS-CIFAR10~\cite{li2017cifar10}), and sequential inputs. It also integrates sequential classification tasks to assess temporal modeling capabilities. For dense prediction tasks, we provide plug-and-play support for SNN-adapted segmentation and detection models, such as FCN~\cite{long2015fully} and FPN~\cite{lin2017feature}, based on MMSeg and MMDet toolchains. 3D point cloud and DVS video detection(PKU-DAVIS-SOD~\cite{li2023sodformer}) are also supported in the latest version. However, since these datasets are not compatible with most models, we do not conduct unified evaluations on them. Some experimental results can be found in the Appendix~\ref{sec:Complicated Datasets}.

\subsection{Multi-task Adaptation}

While early Spiking Transformers (e.g., Spikformer~\cite{zhou2022spikformer}, TIM~\cite{shen2024tim}) were largely limited to classification, recent efforts such as Spike-Driven Transformer V2 have expanded their scope to include dense vision tasks. Our benchmark extends this trajectory by enabling flexible configuration across classification, segmentation, and detection pipelines within a unified framework.

\subsection{Backend-Agnostic Integration}

To enhance accessibility and reusability, our benchmark supports multiple backends such as SpikingJelly~\cite{fang2023spikingjelly}, BrainCog~\cite{zeng2023braincog}, and BrainPy. This backend-agnostic design ensures broad compatibility and enables cross-framework reproducibility. Overall, the framework is robust, extensible, and task-agnostic, offering a solid foundation for developing and evaluating Spiking Transformer architectures.

\section{Experiment}
\label{sec:expirements}
% In benchmarking work, the accuracy and reliability of experimental results are of paramount importance. This section provides a detailed description of the experimental design and presents the results obtained. The primary objective is to demonstrate the performance of several representative Spiking Transformer models across multiple datasets, followed by a comprehensive analysis of the results.
To ensure fair and reliable evaluation, we reproduce several representative Spiking Transformer models under a unified training setup. This section details the experimental protocol and presents the reproduced results on benchmark datasets.

\subsection{Reproduction}
\begin{table}[h]
  \centering
  \renewcommand{\arraystretch}{1.2} 
\caption{Reproduced top-1 accuracy (\%) of Spiking Transformer models on CIFAR-10 and CIFAR-100. *: SGLFormer uses a reduced batch size (16) due to high memory demand. **: SpikingResformer was originally trained with transfer learning; we instead use end-to-end training.}
  \resizebox{0.9\textwidth}{!}{
  \begin{tabular}{c c c c c c}
  \toprule
  \textbf{Model}           & \textbf{Batch-Size} & \textbf{Step} & \textbf{Epoch} & \textbf{CIFAR10 (Acc@1)} &
  \textbf{CIFAR100 (Acc@1)} \\ 
  \midrule
  Spikformer~\cite{zhou2022spikformer}              & 128                & 4             & 400             & 95.12\ (95.51)              & 77.37\ (78.21)               \\ 
  SDT~\cite{yao2023spike}                      & 128                 & 4             & 400             & 95.77\ (95.60)              & 78.29\ (78.40)               \\
  QKFormer~\cite{zhou2024qkformer}                 & 128                 & 4             & 400             & 96.24\ (96.18)              & 79.72\ (81.15)              \\ 
  Spikingformer~\cite{zhou2023spikingformer}            & 128                & 4             & 400             & 95.53\ (95.81)              & 79.12\ (79.21)               \\ 
  Spikformer + SEMM~\cite{zhou2024spikformer}        & 128                & 4             & 400             & 94.98\ (95.78)              & 77.59\ (79.04)               \\ 
  Spiking Wavelet~\cite{fang2024spiking}          & 128                & 4             & 400             & 95.31\ (96.10)              & 76.99\ (79.30)               \\ 
  SGLFormer~\cite{zhang2024sglformer}$^*$              & 16                 & 4             & 400             & 95.88\ (96.76)              & 80.61\ (82.26)               \\ 
  SpikingResformer~\cite{zhou2023spikingformer}$^{**}$         & 128                & 4             & 400             & 95.69\ (97.40)              & 79.45\ (85.98)               \\ 
  \bottomrule
\end{tabular}
}
  \label{tab:reproduced_cifar}
\end{table}

Tab.~\ref{tab:reproduced_cifar} presents our reproduced results on CIFAR-10 and CIFAR-100~\cite{krizhevsky2009learning}. All models are trained using the same optimizer, learning rate, batch size (unless otherwise constrained), training epochs, and random seed. Experiments are conducted on NVIDIA A100 GPUs with 40GB memory.

Overall, our reproduced results are consistent with the original papers. Some models, such as QKFormer, even outperform their reported results, suggesting strong reproducibility. Discrepancies stem mainly from (i) implementation differences, e.g., SpikingResformer originally uses transfer learning, while our setup employs end-to-end training; and (ii) memory limitations, e.g., SGLFormer requires a smaller batch size. To ensure fairness, we avoid dataset- or model-specific tuning and apply a uniform experimental protocol across all baselines. The metrics commonly used to evaluate the framework and reproduction robustness can be found in the Appendix~\ref{sec:robustness}. 

% In Table~\ref{tab:reproduced_cifar}, we report the performance of several representative open-source Spiking Transformer models reproduced on CIFAR-10/100~\cite{krizhevsky2009learning}. All experiments were conducted on NVIDIA A100 40GB GPUs to ensure sufficient resources and consistent evaluation.

% Using a unified framework with identical settings for the optimizer, learning rate, batch size, and training epochs, our reproduced results are largely comparable to the original papers. Notably, models like QKFormer even outperform their reported baselines, demonstrating the reliability of our reproduction.

% Minor discrepancies arise mainly from two sources: First, implementation gaps—for example, Spiking Wavelet Transformer requires wavelet-specific adaptations, SpikingResformer was originally trained with transfer learning whereas we use end-to-end training, and SGLFormer faces memory constraints preventing batch size alignment. Second, for fairness, we uniformly apply the same experimental setup across all models and datasets, avoiding any model- or dataset-specific tuning.

\subsection{Experiments on More Complex Tasks}

To further evaluate the scalability and task generalization of Spiking Transformer models, we test their performance on ImageNet-1K for large-scale classification, ADE20K for semantic segmentation and COCO for object detection, all of which are significantly more complex than CIFAR-level datasets.

\subsubsection{Classification: \ ImageNet-1K}
For ImageNet-1K~\cite{deng2009imagenet} we evaluate only Spikformer~\cite{zhou2024spikformer} and QKFormer~\cite{zhou2024qkformer}: the former is the seminal Spiking-Transformer baseline, while the latter introduces a hierarchical pyramid and currently delivers SOTA accuracy among SNN-based Transformers. Concentrating our limited GPU budget on these two “end-points” lets us cover the full architectural spectrum without incurring the prohibitive cost of training several similar intermediate models. Because ImageNet-1K is orders of magnitude larger and more complex than CIFAR-10/100—and because Spikformer and QKFormer differ greatly in parameter count and memory footprint—forcing a single batch size and epoch schedule would either overflow A100 GPU memory or demand untenable compute. We therefore keep each model’s published regime (QKFormer: 200 epochs × 32/GPU; Spikformer: 300 epochs × 24/GPU), while unifying every other hyper-parameter under a single script; the differing batch sizes and epoch counts are thus an intentional, resource-aware decision rather than an oversight.

The shortfall in our QKFormer accuracy comes from two choices: we evaluated the compact variant and trained every model with one unified script that omits architecture-specific optimisations. This inevitably costs QKFormer a few points, yet the results still validate our reproduction, and we will extend the same benchmark to the remaining Spiking Transformers on ImageNet.

% For ImageNet-1K~\cite{deng2009imagenet}, we evaluate Spikformer~\cite{zhou2024spikformer} and QKFormer~\cite{zhou2024qkformer}, representing the Spiking-Transformer baseline and the state-of-the-art model with a hierarchical pyramid, respectively. By focusing on these two models, we cover the architectural spectrum without the prohibitive cost of training multiple similar models. Given ImageNet-1K's larger size and complexity compared to CIFAR-10/100, and the significant differences in parameter count and memory footprint between Spikformer and QKFormer, we use different batch sizes and epoch schedules, while unifying other hyperparameters. The varying schedules are intentional for resource optimization. The accuracy shortfall in QKFormer is due to evaluating the compact variant and using a unified script that omits architecture-specific optimizations. Despite this, the results validate our reproduction, and we plan to extend the benchmark to other Spiking Transformers on ImageNet.
\subsubsection{Segmentation:\ ADE20K}

\begin{table}[h]
\centering
\caption{Reproduced performance on ImageNet-1K and ADE20K without pretraining.}
\resizebox{0.9\linewidth}{!}{
\begin{tabular}{c|c c c c|c c c}
\toprule
\multirow{2}{*}{\textbf{Model}} & \multicolumn{4}{c|}{\textbf{ImageNet-1K (Classification)}} & \multicolumn{3}{c}{\textbf{ADE20K (Segmentation)}} \\
 & \textbf{Batch Size} & \textbf{Step} & \textbf{Epochs} & \textbf{Acc@1} & \textbf{aAcc} & \textbf{mIoU} & \textbf{mAcc} \\
\midrule
QKFormer~\cite{zhou2024qkformer}             & 256 & 4 & 200 & 73.88 & -     & -     & -     \\
Spikformer~\cite{zhou2023spikingformer}      & 192 & 4 & 300 & 73.69 & 69.80 & 23.51 & 31.43 \\
Spikformer + SEMM~\cite{zhou2024spikformer}  & -   & - & -   &  -    & 63.41 & 13.13 & 19.76 \\
SDT~\cite{yao2023spike}                      & 384   & 4 & 300   &  71.96    & 63.45 & 12.08 & 17.17 \\
\bottomrule
\end{tabular}
}
\label{tab:complex_tasks}
\end{table}

% \begin{wrapfigure}[15]{Hr}{0.55\textwidth}
%   \centering
%   \includegraphics[width=\linewidth]{images/seg_predict.jpg}
%   \caption{\small Segmentation predictions on ADE20K for three Spiking Transformer variants.}
%   \label{fig: seg_predict}
% \end{wrapfigure}
For semantic segmentation on ADE20K~\cite{zhou2017scene}, only SDT~\cite{yao2023spike} had been previously evaluated. We conduct a fair comparison by retraining Spikformer ~\cite{zhou2024spikformer}and SEMM~\cite{zhou2024spiking} \textbf{without pretraining}. Interestingly, both Spikformer variants outperform SDT under identical settings, despite SDT's original paper reporting strong performance with pretraining. This implies that with appropriate initialization strategies, Spikformer-based models could potentially surpass existing baselines in dense prediction tasks. The segmentation result can be viewed in Fig.~\ref{fig: seg_predict}

These results demonstrate that Spiking Transformers, when carefully trained, are capable of scaling to more complex tasks beyond image classification, including semantic segmentation, and are promising candidates for broader real-world neuromorphic applications.

\subsubsection{Detection:\ COCO}

Object detection requires simultaneous localisation and classification across diverse scales, a challenge naturally addressed by multi-resolution features. Among existing Spiking Transformers, only \textbf{SDTv2} produces genuine multi-scale outputs, making it the sole candidate for COCO. Training SDTv2 from scratch yields poor box regressors, whereas ImageNet pre-training boosts mAP by an order of magnitude (Tab.~\ref{tab:det} \& Fig.~\ref{fig:det}). Unlike segmentation, where models converge without priors, detection proves highly sensitive to object-level cues and foreground–background balance. Thus, effective spiking detectors must combine multi-scale backbones with strong pre-training. These findings 
inspire future spiking designs with built-in pyramids and large-scale (self-supervised) pre-training to bridge the gap with ANNs and enable energy-efficient event-driven detection. More detailed results are reported in Appendix~\ref{ap:det}.

\section{Analysis}
\label{sec:analysis}
Transformers' ability to model sequential dependencies has recently been questioned, particularly regarding the actual benefits of sparse attention in SNN contexts. Among existing models, Spikformer first introduced attention mechanisms into SNNs, while SDT significantly reduced their computational complexity. Many subsequent works, including Spikformer+SEMM (which incorporates a Mixture of Experts with minimal modification), are derived from or inspired by these two. We focus our analysis on these three representative models.

\subsection{Neuron Model Evaluation}

To investigate the impact of different spiking neuron types on model performance, we replace the default LIF neuron with five widely used variants: PLIF~\cite{fang2021incorporating}, CLIF~\cite{huang2024clif}, GLIF~\cite{yao2022glif}, and KLIF~\cite{jiang2023klif}. These models extend the basic LIF neuron~\cite{hunsberger2015spiking} by incorporating enhancements such as learnable time constants, gating mechanisms, and surrogate gradient improvements. To quantify the influence of neuron choice, we replace the default LIF cell with four mainstream variants—PLIF \cite{fang2021incorporating}, CLIF \cite{huang2024clif}, GLIF \cite{yao2022glif}, and KLIF \cite{jiang2023klif}. Each variant augments the canonical LIF formulation~\cite{hunsberger2015spiking} with additional biological or optimization benefits, ranging from a learnable membrane constant (PLIF) to gated internal states (GLIF) and surrogate-gradient refinements (CLIF, KLIF). Detailed 
explaination can be found in Appendix~\ref{ap:neuron}.

\begin{figure}[t]
  \centering
  \begin{minipage}[c]{0.45\textwidth} % 图：垂直居中对齐
    \centering
  \includegraphics[width=\linewidth]{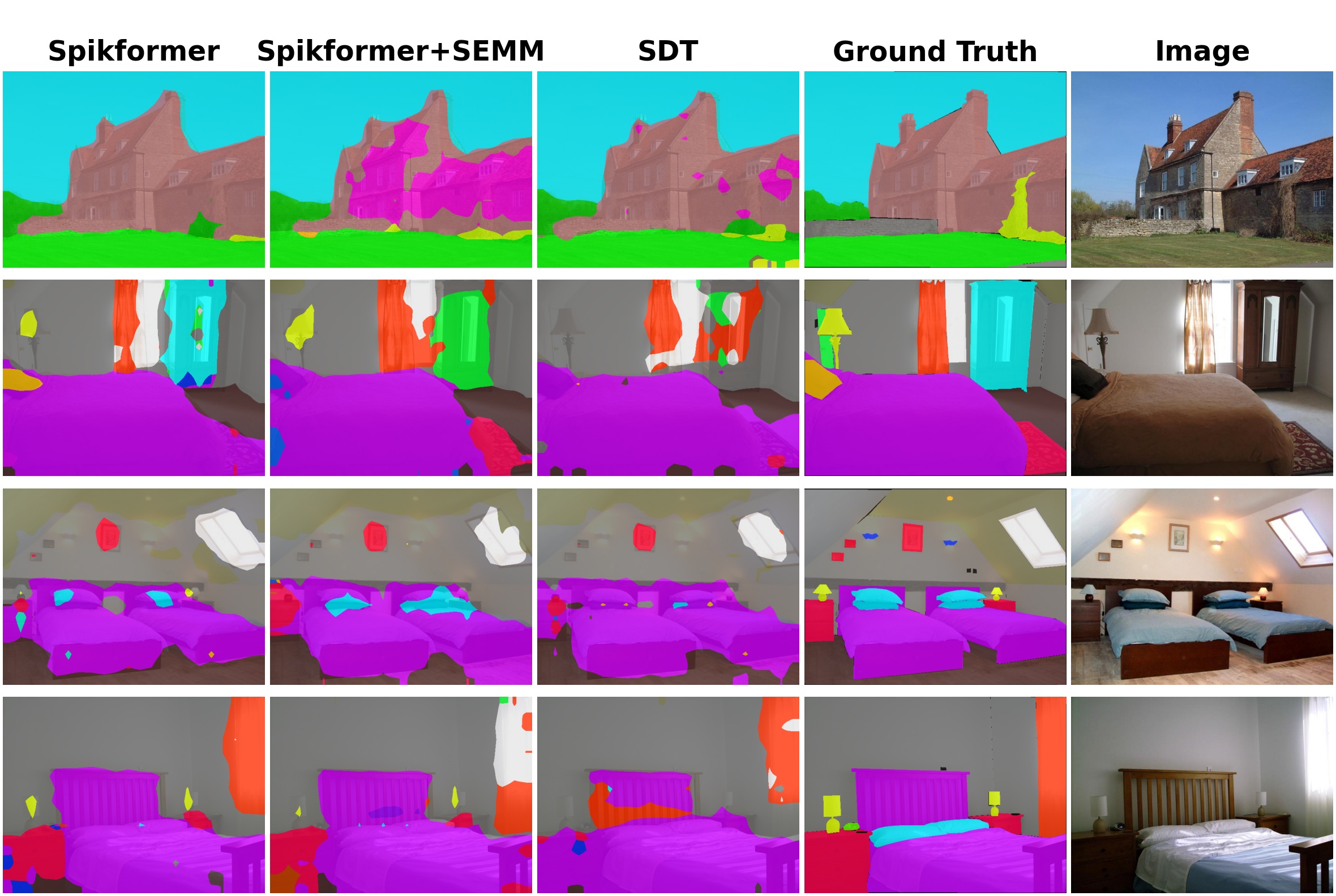}
  \caption{\small Segmentation predictions on ADE20K for three Spiking Transformer variants.}
  \label{fig: seg_predict}
  \end{minipage}
  \hfill
  \begin{minipage}[c]{0.47\textwidth} % 右侧表格
  
    \centering
  \captionof{table}{\small Top-1 accuracy (\%) of different neuron types on CIFAR-10.}
  \resizebox{\linewidth}{!}{
  \begin{tabular}{c c c c c c}
      \toprule
      \textbf{Model} & \textbf{LIF} & \textbf{CLIF} & \textbf{GLIF} & \textbf{KLIF} & \textbf{PLIF} \\
      \midrule
      Spikformer~\cite{zhou2022spikformer}          & 95.12 & 95.38 & 95.41 & 95.85 & \textbf{96.06}  \\
      SDT~\cite{yao2023spike}                       & 95.77 & 95.49 & 95.45 & 95.63 & \textbf{95.91}  \\
      Spikformer + SEMM~\cite{zhou2024spikformer}   & 94.98 & 95.44 & \textbf{95.78} & 95.59 & 95.66  \\
      \bottomrule
  \end{tabular}
  \label{tab:node}
  }
    \centering
    \captionof{table}{\small SDTv2 detection result on COCO. Step:1; \ Epoch:10.}
    \resizebox{0.85\textwidth}{!}{
    \begin{tabular}{c c c}
  \toprule
  \textbf{Pre-training} &  \textbf{bbox mAP@0.5} & \textbf{segm mAP@0.5} \\ 
  \midrule
  \textsc{No}                         & 1.7             & 1.6              \\ 
 \textsc{Yes}                              & 10.5             & 10.4              \\
  \bottomrule
\end{tabular}
      \label{tab:det}
}
  \end{minipage}
\end{figure}

% To quantify the influence of neuron choice, we replace the default LIF cell with four mainstream variants—PLIF \cite{fang2021incorporating}, CLIF \cite{huang2024clif}, GLIF \cite{yao2022glif}, and KLIF \cite{jiang2023klif}. Each augments the canonical LIF formulation \cite{hunsberger2015spiking} with additional biological or optimisation benefits, ranging from a learnable membrane constant (PLIF) to gated internal states (GLIF) and surrogate-gradient refinements (CLIF, KLIF). Formal information appear in Appendix~\ref{ap:neuron}.

Tab.~\ref{tab:node} reports consistent accuracy gains across three backbone architectures once these enhanced neurons are introduced. PLIF delivers the largest improvement—surpassing even architectural upgrades on Spikformer—yet it adds only one scalar parameter. We attribute the gain to richer, more biologically plausible membrane dynamics that encourage sparse, spike-driven learning.

These results indicate that Spiking Transformers lean more on intrinsic neuron dynamics than on explicit temporal modules. Progress therefore calls for biologically faithful yet efficient additions—such as dendritic processing or multi-compartment cells—perhaps embedded in hybrid recurrent-spiking frameworks.

\subsection{Sequence Modeling}

Recent studies~\cite{stan2024learning} question the ability of standard Transformers to model long-range temporal dependencies, prompting alternatives like Spiking SSM~\cite{shen2025spikingssms}. Datasets such as sCIFAR~\cite{chang2017dilatedrecurrentneuralnetworks} and (p)sMNIST~\cite{le2015simplewayinitializerecurrent} serialize 2D images into 1D sequences, emphasizing temporal structure. Spiking SSM processes inputs at the pixel level (e.g., 784 steps for a full MNIST image), incurring high computational costs. To adapt Spiking Transformers for serialized inputs, we replace 2D convolutions in the SPS module with 1D convolutions.

\begin{table}[h]
  \centering
  \renewcommand{\arraystretch}{1} 
  \caption{Top-1 accuracy (\%) of selected models on sequential image classification datasets. Batch size = 128, epochs = 400, steps = 4.\ *: Original ViT; \  **: ViT with 4-layer-conv embedding.}
  \resizebox{0.65\linewidth}{!}{
  \begin{tabular}{c c c c c}
      \toprule
      \textbf{Model} & \textbf{SNN} & \textbf{sMNIST} & \textbf{psMNIST} & \textbf{sCIFAR} \\
      \midrule
      FlexTCN~\cite{romero2021flexconv}             & \textsc{No}  & 99.62 & 98.63 & 80.82 \\
      SMPConv~\cite{kim2023smpconv}                 & \textsc{No}  & \textbf{99.75} & \textbf{99.10} & 84.86 \\
      LMUformer~\cite{liu2024lmuformer}            & \textsc{No} & -     & 98.55 & -     \\
      ViT~\cite{dosovitskiy2020image}  $^*$             & \textsc{No}  & 98.00 & 97.73 & 74.95 \\
      ViT\cite{dosovitskiy2020image}        + SPS     $^{**}$                                 & \textsc{No}  & 99.19 & 98.19 & \textbf{85.62} \\
      \midrule
      SpikingSSM~\cite{shen2025spikingssms}        & \textsc{Yes}& 99.60 & 98.40 & -     \\
      SpikingLMUformer~\cite{liu2024lmuformer}     & \textsc{Yes}& -     & 97.92 & -     \\
      \midrule
      Spikformer~\cite{zhou2022spikformer}         & \textsc{Yes} & 98.84 & 97.97 & 84.26 \\
      SDT~\cite{yao2023spike}                      & \textsc{Yes}& 98.77 & 97.80 & 82.31 \\
      Spikformer + SEMM~\cite{zhou2024spikformer}  & \textsc{Yes} & 99.33 & 98.46 & 85.61 \\
      \bottomrule
  \end{tabular}
  }
  \label{tab:Sequence}
\end{table}
As shown in Tab.~\ref{tab:Sequence}, SNN-based Spiking Transformers lag behind ANN counterparts like ViT+SPS and SMPConv, even with MoE enhancements in Spikformer+SEMM. This suggests that spike-based attention mechanisms are more suited for spatial rather than temporal modeling.
We hypothesize that performance limitations stem from the restricted number of training steps and sparse neuron activations, which weaken temporal expressiveness. Future work should explore spatiotemporal attention designs and biologically inspired mechanisms like spike-timing-dependent plasticity (STDP) to improve temporal modeling without excessive computational cost.

\subsection{Encoding Schemes}
\label{para:Ecoding_analysis}
% \begin{wraptable}[9]{r}{0.55\columnwidth}
%   \centering
%   \captionof{table}{\small Top-1 accuracy (\%) of different encoding methods on CIFAR-10. Batch size: 128, Epoch: 400, Step: 4.}
%   \renewcommand{\arraystretch}{1}
%   \resizebox{1\linewidth}{!}{
%   \begin{tabular}{c c c c c}
%       \toprule
%       \textbf{Model}  & \textbf{Direct} & \textbf{Phase} & \textbf{Rate} & \textbf{TTFS} \\
%       \midrule
%       Spikformer~\cite{zhou2022spikformer}          & \textbf{95.12} & 82.75 & 82.83 & 82.10 \\
%       SDT~\cite{yao2023spike}                        & \textbf{95.77} & 85.37 & 83.77 & 84.30 \\
%       Spikformer + SEMM~\cite{zhou2024spikformer}    & \textbf{94.98} & 85.81 & 83.04 & 83.37 \\
%       \bottomrule
%   \end{tabular}
%   }
%   \label{tab:encoding_test}
% \end{wraptable}
% RGB inputs can be converted to spikes using four common encoding schemes: direct, phase, rate, and TTFS~\cite{adrian1926impulses, park2020t2fsnn, kim2018deep}. As shown in Tab.~\ref{tab:encoding_test}, direct encoding, a lossless approach ensures that no information is lost—which repeats the entire image at every time step matches current Spiking Transformers that compute attention independently across steps, and therefore yields the highest Top-1 accuracy on all models. In contrast, the sparser phase, rate, and TTFS encodings reduce spike density and weaken spatial coherence, leading to lower accuracy and highlighting the need for future architectures with temporally aware attention or recurrent mechanisms.
RGB inputs can be converted to spikes through four encoding schemes: direct, phase, rate, and TTFS~\cite{adrian1926impulses, park2020t2fsnn, kim2018deep}. As shown in Tab.~\ref{tab:encoding_test}, direct encoding—being lossless and repeating the full image at each timestep—aligns with current Spiking Transformers that compute attention independently, yielding the highest Top-1 accuracy. In contrast, the sparser phase, rate, and TTFS encodings reduce spike density and spatial coherence, leading to lower accuracy and emphasizing the need for temporally aware attention or recurrent designs.

\subsection{Sparse Attention Analysis}

In Sec.~\ref{para:Ecoding_analysis}, we observed that Spiking Transformers struggle to model temporal dependencies. Here, we further examine whether their attention mechanisms meaningfully contribute to spatial feature extraction.

\paragraph{Randomized Attention.}  
To ablate the role of attention, we fix the $Q$ and $K$ branches to randomly initialized, frozen weights, while keeping $V$ trainable for gradient propagation:
\begin{gather}
Q = \text{LIF}(W^Q_{\text{detach}} X),\quad K = \text{LIF}(W^K_{\text{detach}} X),\quad V = \text{LIF}(W^V X)
\label{eq:random_attn}
\end{gather}

We apply this to three representative models (Spikformer, SDT, and Spikformer+SEMM), and include ViT as an ANN-based baseline. As shown in Fig.~\ref{tab:random_attn}, Spiking Transformers maintain performance under randomized attention (drop < 0.35\%), with Spikformer+SEMM even slightly improving. In contrast, ViT suffers a notable drop, indicating its strong reliance on attention.

% \begin{figure}[t]
%   \centering
%   \begin{minipage}[c]{0.48\textwidth} % 图：垂直居中对齐
%     \centering
%     \includegraphics[width=\linewidth]{images/random_attn.pdf}
%     \caption{\small Impact of randomized attention and reduced SPS depth}
%     \label{fig:random_attn}
%   \end{minipage}
%   \hfill
%   \begin{minipage}[c]{0.48\textwidth} % 表：垂直居中对齐
%     \centering
%     \renewcommand{\arraystretch}{1}
%     \captionof{table}{\small Top-1 accuracy (\%) with SDSA-v3 under varying SPS depths.}
%     \resizebox{1\linewidth}{!}{
%       \begin{tabular}{c c c c}
%         \toprule
%         \textbf{Model} & \textbf{SPS (4 conv)} & \textbf{SPS-2conv} & \textbf{SPS-1conv} \\
%         \midrule
%         Spikformer~\cite{zhou2022spikformer} & 95.57 & 93.43 & 89.97 \\
%         SDT~\cite{yao2023spike}              & 96.38 & 94.68 & 87.33 \\
%         Spikformer + SEMM~\cite{zhou2024spikformer} & 95.83 & 93.37 & 84.95 \\
%         \bottomrule
%       \end{tabular}
%       \vspace{2pt}
%     }
%     \label{tab:sdsa-v3}
%   \end{minipage}
% \end{figure}

\begin{figure}[t]
  \centering
  \begin{minipage}[c]{0.45\textwidth} % 图：垂直居中对齐
    % \centering
    % \includegraphics[width=\linewidth]{images/random_attn.pdf}
    % \caption{\small Impact of randomized attention and reduced SPS depth}
    % \label{fig:random_attn}
    \centering
    \includegraphics[width=\linewidth]{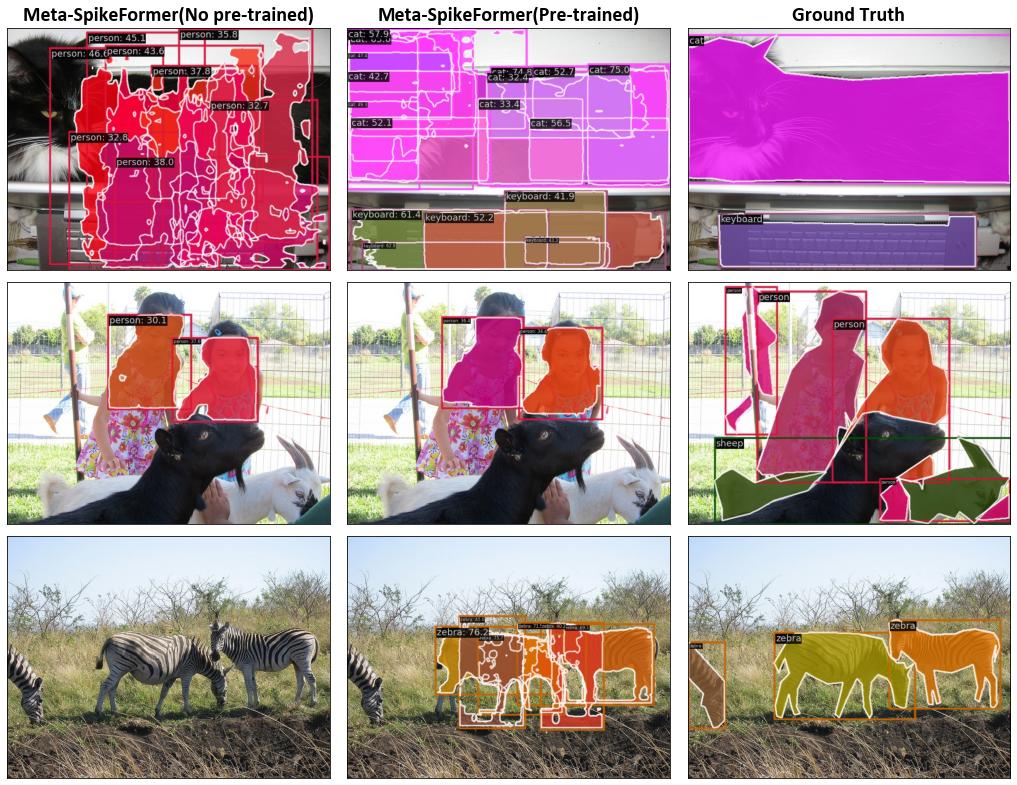}
    \caption{\small Result of SDTv2 on COCO datasets.}
    \label{fig:det}

  \end{minipage}
  \hfill
  \begin{minipage}[c]{0.47\textwidth} % 右侧表格
  
    \centering
    \captionof{table}{\small Top-1 accuracy (\%) of different encoding methods on CIFAR-10. Batch size: 128, Epoch: 400, Step: 4.}
    \resizebox{1\linewidth}{!}{
      \begin{tabular}{c c c c c}
          \toprule
          \textbf{Model}  & \textbf{Direct} & \textbf{Phase} & \textbf{Rate} & \textbf{TTFS} \\
          \midrule
          Spikformer~\cite{zhou2022spikformer}          & \textbf{95.12} & 82.75 & 82.83 & 82.10 \\
          SDT~\cite{yao2023spike}                        & \textbf{95.77} & 85.37 & 83.77 & 84.30 \\
          Spikformer + SEMM~\cite{zhou2024spikformer}    & \textbf{94.98} & 85.81 & 83.04 & 83.37 \\
          \bottomrule
            \label{tab:encoding_test}
      \end{tabular}
    }
    \centering
    \renewcommand{\arraystretch}{1}
    \captionof{table}{\small Top-1 accuracy (\%) with SDSA-v3 under varying SPS depths.}
    \resizebox{1\linewidth}{!}{
      \begin{tabular}{c c c c}
        \toprule
        \textbf{Model} & \textbf{SPS (4 conv)} & \textbf{SPS-2conv} & \textbf{SPS-1conv} \\
        \midrule
        Spikformer~\cite{zhou2022spikformer} & 95.57 & 93.43 & 89.97 \\
        SDT~\cite{yao2023spike}              & 96.38 & 94.68 & 87.33 \\
        Spikformer + SEMM~\cite{zhou2024spikformer} & 95.83 & 93.37 & 84.95 \\
        \bottomrule
        \label{tab:sdsa-v3}
      \end{tabular}
    }
  \end{minipage}
\end{figure}

% \begin{wrapfigure}[15]{Hr}{0.4\textwidth} 
%   % [16] 所需要的行高
%   \centering
%   \vspace{-2pt}  % 上方收紧，可按需调整
%   \includegraphics[width=0.4\textwidth]{images/random_attn.pdf}
%   \caption{\small Impact of randomized attention and reduced SPS depth}
%   \label{fig:random_attn}
%   \vspace{-8pt}  % 下方收紧，可按需调整
% \end{wrapfigure}

\paragraph{Reduced Convolutional Depth in SPS.}  
\begin{wraptable}[9]{r}{0.56\textwidth}
  \small
  \centering
  \caption{\small Comparison of Acc@1 for Different Model Configurations on CIFAR-10.}
  \label{tab:random_attn}
  \resizebox{1\linewidth}{!}{
  \begin{tabular}{c c c c c}
    \toprule
    \textbf{Model} & \textbf{Original }& \textbf{Random\_Attn} & \textbf{SPS (1 Conv)} & \textbf{SPS (2 Conv)} \\
    \midrule
    Spikformer             & 95.12 & 94.96 & 78.21 & 91.92 \\
    SDT                    & 95.77 & 95.45 & 77.34 & 94.03 \\
    Spikformer+SEMM        & 94.98 & 95.57 & 89.24 & 93.33 \\
    \midrule
    ANN\_ViT               & 90.89 & 88.46 & —     & —     \\
    \bottomrule
  \end{tabular}
  }
\end{wraptable}
We next evaluate model robustness under reduced convolutional depth in the SPS module. When decreasing SPS from four to two and one layers, performance deteriorates sharply across all models. With only one conv layer, models behave like pure attention-based Spiking Transformers and fail to match baseline accuracy—highlighting the dominant role of convolution in feature extraction.

\paragraph{Replacement with SDSA-v3.}  
To test whether stronger attention can compensate for weaker convolutional backbones, we replace SSA with SDSA-v3~\cite{yao2025scaling,chen2018encoder}, where QKV are generated using depthwise separable convolutions:
\begin{equation}
\begin{gathered}
W  = 
\textbf{SSA+(SEMM)}:  \text{Linear}(\cdot); \ 
\textbf{SDSA}: \text{ConvBN}(\cdot);\ 
\textbf{SDSA-V3}:  \text{BN}(\text{SepConv}(\cdot))
\label{Eq:sdsa-v3}
\end{gathered}
\end{equation}

% \begin{wraptable}[8]{r}{0.55\columnwidth}
%   \vspace{-4pt}
%   \centering
%   \caption{\small Top-1 accuracy (\%) with SDSA-v3 under varying SPS depth. Batch size: 128, Epoch: 400, Step: 4.}
%   \setlength{\tabcolsep}{4pt}
%   \renewcommand{\arraystretch}{1}
%   \resizebox{1\linewidth}{!}{
%     \begin{tabular}{c c c c}
%       \toprule
%       \textbf{Model} &  \textbf{SPS (4 conv)} & \textbf{SPS-2conv} & \textbf{SPS-1conv} \\
%       \midrule
%       Spikformer~\cite{zhou2022spikformer}             & 95.57 & 93.43 & 89.97 \\
%       SDT~\cite{yao2023spike}                          & 96.38 & 94.68 & 87.33 \\
%       Spikformer + SEMM~\cite{zhou2024spikformer}       & 95.83 & 93.37 & 84.95 \\
%       \bottomrule
%   \end{tabular}
%   }
%   \label{tab:sdsa-v3}
%   \vspace{-6pt}
% \end{wraptable}

Even with SDSA-v3, performance remains positively correlated with convolutional depth (Tab.~\ref{tab:sdsa-v3}). While SDSA-v3 reduces the performance gap, it does not eliminate reliance on convolution. These findings suggest that current spike-based attention mechanisms contribute limited spatial modeling capacity, with most representational power still residing in the convolutional frontend.

\subsection{Energy Efficiency Modeling}

Energy modeling in SNNs traditionally estimates cost based on the number of accumulate (AC) operations, whereas for ANNs, it relies on multiply-accumulate (MAC) operations.
However, we argue that current methodologies overlook two critical aspects:

\begin{itemize}
    \item \textbf{Quantized ANNs are underestimated in efficiency.} Bit-serial execution in low-bitwidth ANNs~\cite{eckert2018neural,zhao2020bitpruner} can transform MACs into sequences of ACs, which can exploit bit-level sparsity to skip ineffectual operations—similar to spike sparsity in SNNs. This makes quantized ANNs significantly more efficient than previously assumed.
    
    \item \textbf{Memory access energy is often ignored.} Previous comparisons often overlook the energy cost associated with on-chip and off-chip memory accesses. In SNNs, high-precision membrane potentials must be maintained and updated throughout multiple time steps, necessitating frequent accesses. In contrast, ANNs only require writing back quantized activations, which has less memory burden. This omission in existing energy models can result in an overestimation of the energy efficiency of SNNs relative to ANNs.
\end{itemize}

To address these gaps, we propose a analytical framework that models both spiking and quantized neural networks shown in Tab.~\ref{tab:energy-model}, and Tab.~\ref{tab:energy-breakdown} presents an quantitive comparison. While the spiking transformer show a small advantage in compute efficiency over the quantized transformer, its overall energy consumption is unexpectedly higher once memory access is factored in.

\begin{table}[h!]
  \centering
  \caption{\small Energy analysis modeling. $F_{Conv}$ and $F_{Mlp}$ denote FLOPs of Conv and MLP modules in ANNs. $B$ is the quantized bit-width in quantized Transformers; $T$ is the time steps in spiking Transformers. $R_s$ (firing rate) and $R_b$ (bit rate) represent spike sparsity and bit-level sparsity of the quantized activation. $E_{Mac}=4.6pJ$, $E_{Ac}=0.9pJ$, and $E_{Mem}=3.12pJ$ denote energy per MAC, AC, and memory access (per bit energy access from a 16MB cache), respectively~\cite{horowitz20141}.}
  \vspace{0.3em} % 根据需要调整标题与表格的距离
  \setlength{\tabcolsep}{4pt}   % 列间距
  \renewcommand{\arraystretch}{1.2} % 行高
  %———整体缩放 0.85× ———
  \resizebox{0.8\linewidth}{!}{
  \begin{tabular}{c c c c c c}
    \toprule
    \textbf{Module} & \textbf{Op.} & \textbf{Type} &
      \multirow{2}{*}{\shortstack{\textbf{Vanilla}\\\textbf{Transformer}}} &
      \multirow{2}{*}{\shortstack{\textbf{Quantized}\\\textbf{Transformer}}} &
      \multirow{2}{*}{\shortstack{\textbf{Spiking}\\\textbf{Transformer}}} \\
    & & & & & \\
    \midrule
    \multirow{2}{*}{SPS}
      & \multirow{2}{*}{Conv}
      & Compute & $E_{Mac}F_{Conv}$ & $B R_b\cdot E_{Ac}F_{Conv}$
                 & $T R_s\cdot E_{Ac}F_{Conv}$ \\
      &         & Memory  & $32\cdot E_{Mem} C_o HW$
                 & $B\cdot E_{Mem} C_o HW$
                 & $32T\cdot E_{Mem} C_o HW$ \\
    \midrule
    \multirow{6}{*}{\shortstack{Self\\Attention}}
      & \multirow{2}{*}{Q,K,V}
      & Compute & $E_{Mac}3ND^2$
                 & $B R_b\cdot E_{Ac}3ND^2$
                 & $T R_s\cdot E_{Ac}3ND^2$ \\
      &         & Memory  & $32\cdot E_{Mem} 3ND$
                 & $B\cdot E_{Mem} 3ND$
                 & $32T\cdot E_{Mem} 3ND$ \\
      & \multirow{2}{*}{$f$(Q,K,V)}
      & Compute & $E_{Mac}2N^2D$
                 & $B R_b\cdot E_{Ac}2N^2D$
                 & $T R_s\cdot E_{Ac} ND$ \\
      &         & Memory  & $32\cdot E_{Mem} 2N^2$
                 & $B\cdot E_{Mem} 2N^2$
                 & $32T\cdot E_{Mem} ND$ \\
      & \multirow{2}{*}{Linear}
      & Compute & $E_{Mac}F_{Mlp}$
                 & $B R_b\cdot E_{Ac}F_{Mlp}$
                 & $T R_s\cdot E_{Ac}F_{Mlp}$ \\
      &         & Memory  & $32\cdot E_{Mem} C_o$
                 & $B\cdot E_{Mem} C_o$
                 & $32T\cdot E_{Mem} C_o$ \\
    \midrule
    \multirow{2}{*}{MLP}
      & \multirow{2}{*}{Linear}
      & Compute & $E_{Mac}F_{Mlp}$
                 & $B R_b\cdot E_{Ac}F_{Mlp}$
                 & $T R_s\cdot E_{Ac}F_{Mlp}$ \\
      &         & Memory  & $32\cdot E_{Mem} C_o$
                 & $B\cdot E_{Mem} C_o$
                 & $32T\cdot E_{Mem} C_o$ \\
    \bottomrule
  \end{tabular}}
  \label{tab:energy-model}
\end{table}

% Tab.~\ref{tab:energy-breakdown} presents a comparison of energy consumption—covering both computation and memory access—among the vanilla transformer, the quantized transformer, and the spiking transformer. While the spiking transformer show a small advantage in compute efficiency over the quantized transformer, its overall energy consumption is unexpectedly higher once memory access is factored in.

\begin{table}[h!]
  \centering
  \caption{\small Energy analysis comparison.}
  \renewcommand{\arraystretch}{1.2}
    \resizebox{0.7\linewidth}{!}{
  \begin{tabular}{l c c c c c}
    \toprule
    \textbf{Model} & \textbf{Param} & \textbf{Neuron} & \textbf{Compute} & \textbf{Mem} & \textbf{Total} \\
    \midrule
    Transformer-8-512 Float  & 29.68M & 14M & 41.77mJ & 1.39mJ & 43.16mJ \\
    Transformer-8-512 Quant  & 29.68M & 14M & 16.34mJ & 0.17mJ & 16.51mJ \\
    SpikingTransformer-8-512 & 29.68M & 14M & 11.57mJ & 5.59mJ & 17.16mJ \\
    \bottomrule
  \end{tabular}}
  \label{tab:energy-breakdown}
\end{table}

\section{Future Work}

While recent progress in Spiking Transformers has mainly aimed to boost task performance, our results indicate that directly transplanting ANN modules like attention or convolution overlooks key SNN principles. Future work should move beyond performance-oriented adaptation and draw from neuroscience, exploring mechanisms such as dendritic computation, STDP, and temporal coding to design spike-native architectures that are more efficient, robust, and interpretable.

% \section{Future Work}
% Spiking Transformer is a promising and continually evolving field. Based on our aforementioned reproduction, experiments, and analysis, we believe future work should proceed in the following directions:
% \begin{itemize}
%     \item Maintain the benchmark by tracking and updating the latest releases of Spiking Transformers.
%     \item Design Spiking Transformer architectures with stronger temporal information utilization capabilities, and develop attention mechanisms more suited for SNNs to reduce the heavy reliance on convolutions.
%     \item We have already established the framework for \textbf{One Model for Multi Tasks}, and we aim to design a multi-task adaptable Spiking Transformer as the foundational paradigm for the future.
% \end{itemize}
\section{Conclusion}
In this work, we present STEP, a unified benchmarking framework for Spiking Transformers, aiming to standardize evaluation across architectures, datasets, and tasks. STEP integrates diverse implementations under a consistent pipeline, supporting classification, segmentation, and detection on both static and event-based datasets. Through extensive experiments, we reproduced and compared multiple representative models, revealing that current Spiking Transformers rely heavily on convolutional preprocessing while benefiting only marginally from attention mechanisms. Our module-wise ablation further demonstrates that the choice of spiking neuron model and input encoding has a non-trivial impact on final performance, highlighting the importance of biologically inspired design. We also revisited energy efficiency comparisons between SNNs and ANNs. By introducing a unified analytical model that incorporates compute sparsity, bitwidth effects, and memory access costs, we showed that quantized ANNs may be more competitive than previously assumed, urging more careful benchmarking. Taken together, our study highlights the need for deeper integration of neuroscience principles and task-aligned architectural innovations. We hope STEP can serve as a foundation for building truly spike-native Transformers that are efficient, robust, and biologically grounded.

\newpage
\section*{Acknowledgement}
This work is supported by the National Natural Science Foundation of China (Grant No. 62406325).
\bibliographystyle{unsrt}
\bibliography{neurips_2025}

\section*{NeurIPS Paper Checklist}

\begin{enumerate}

\item {\bf Claims}
    \item[] Question: Do the main claims made in the abstract and introduction accurately reflect the paper's contributions and scope?
    \item[] Answer: \answerYes{} % Replace by \answerYes{}, \answerNo{}, or \answerNA{}.
    \item[] Justification: Our abstract and introduction are centered around two key contributions: (1) We propose a unified framework for Spiking Transformers, designed to facilitate the development and evaluation of models across multiple backbones and tasks; (2) Built upon this framework, we conduct a systematic assessment of representative Spiking Transformer models, revealing several objective limitations. Our findings offer constructive insights and practical guidelines for future advancements in Spiking Transformer research.
    \item[] Guidelines:
    \begin{itemize}
        \item The answer NA means that the abstract and introduction do not include the claims made in the paper.
        \item The abstract and/or introduction should clearly state the claims made, including the contributions made in the paper and important assumptions and limitations. A No or NA answer to this question will not be perceived well by the reviewers. 
        \item The claims made should match theoretical and experimental results, and reflect how much the results can be expected to generalize to other settings. 
        \item It is fine to include aspirational goals as motivation as long as it is clear that these goals are not attained by the paper. 
    \end{itemize}

\item {\bf Limitations}
    \item[] Question: Does the paper discuss the limitations of the work performed by the authors?
    \item[] Answer: \answerYes{}% Replace by \answerYes{}, \answerNo{}, or .
    \item[] Justification: Owing to limitations in time and computational resources, we refrain from exhaustively evaluating every existing Spiking Transformer and instead concentrate on a subset that best represents the current architectural landscape. Certain models are intrinsically ill-suited to specific downstream tasks—for instance, Spikformer lacks the structural components required for object detection—so we cannot reproduce their performance on those benchmarks. Moreover, as a benchmark framework, fairness dictates that every model be trained and tested under an identical experimental setup; thus, task-specific tricks reported in the original papers are intentionally omitted. These methodological choices, while essential for consistency, inevitably lead to modest discrepancies between our reproduced results and the figures originally published.
    \item[] Guidelines:
    \begin{itemize}
        \item The answer NA means that the paper has no limitation while the answer No means that the paper has limitations, but those are not discussed in the paper. 
        \item The authors are encouraged to create a separate "Limitations" section in their paper.
        \item The paper should point out any strong assumptions and how robust the results are to violations of these assumptions (e.g., independence assumptions, noiseless settings, model well-specification, asymptotic approximations only holding locally). The authors should reflect on how these assumptions might be violated in practice and what the implications would be.
        \item The authors should reflect on the scope of the claims made, e.g., if the approach was only tested on a few datasets or with a few runs. In general, empirical results often depend on implicit assumptions, which should be articulated.
        \item The authors should reflect on the factors that influence the performance of the approach. For example, a facial recognition algorithm may perform poorly when image resolution is low or images are taken in low lighting. Or a speech-to-text system might not be used reliably to provide closed captions for online lectures because it fails to handle technical jargon.
        \item The authors should discuss the computational efficiency of the proposed algorithms and how they scale with dataset size.
        \item If applicable, the authors should discuss possible limitations of their approach to address problems of privacy and fairness.
        \item While the authors might fear that complete honesty about limitations might be used by reviewers as grounds for rejection, a worse outcome might be that reviewers discover limitations that aren't acknowledged in the paper. The authors should use their best judgment and recognize that individual actions in favor of transparency play an important role in developing norms that preserve the integrity of the community. Reviewers will be specifically instructed to not penalize honesty concerning limitations.
    \end{itemize}

\item {\bf Theory assumptions and proofs}
    \item[] Question: For each theoretical result, does the paper provide the full set of assumptions and a complete (and correct) proof?
    \item[] Answer: \answerNA{} % Replace by \answerYes{}, \answerNo{}, or \answerNA{}.
    \item[] Justification: This work primarily presents a comprehensive benchmark. It is grounded in extensive empirical experiments and observations, with minimal reliance on theoretical derivations.
    \item[] Guidelines:
    \begin{itemize}
        \item The answer NA means that the paper does not include theoretical results. 
        \item All the theorems, formulas, and proofs in the paper should be numbered and cross-referenced.
        \item All assumptions should be clearly stated or referenced in the statement of any theorems.
        \item The proofs can either appear in the main paper or the supplemental material, but if they appear in the supplemental material, the authors are encouraged to provide a short proof sketch to provide intuition. 
        \item Inversely, any informal proof provided in the core of the paper should be complemented by formal proofs provided in appendix or supplemental material.
        \item Theorems and Lemmas that the proof relies upon should be properly referenced. 
    \end{itemize}

    \item {\bf Experimental result reproducibility}
    \item[] Question: Does the paper fully disclose all the information needed to reproduce the main experimental results of the paper to the extent that it affects the main claims and/or conclusions of the paper (regardless of whether the code and data are provided or not)?
    \item[] Answer: \answerYes{} % Replace by \answerYes{}, \answerNo{}, or \answerNA{}.
    \item[] Justification: This work is primarily built upon the BrainCog platform, where we develop a unified Spiking Transformer framework that supports a wide range of tasks, including classification, detection, and segmentation. Detailed experimental results are provided in both the main text and the appendix. As the Dataset \& Benchmark Track follows a single-blind review policy, all code and experiments have been made publicly available; the link can be found in the abstract. Key experimental hyperparameters are reported in the paper. For full details, the complete set of configurations is available in the codebase's configuration files.

    \item[] Guidelines:
    \begin{itemize}
        \item The answer NA means that the paper does not include experiments.
        \item If the paper includes experiments, a No answer to this question will not be perceived well by the reviewers: Making the paper reproducible is important, regardless of whether the code and data are provided or not.
        \item If the contribution is a dataset and/or model, the authors should describe the steps taken to make their results reproducible or verifiable. 
        \item Depending on the contribution, reproducibility can be accomplished in various ways. For example, if the contribution is a novel architecture, describing the architecture fully might suffice, or if the contribution is a specific model and empirical evaluation, it may be necessary to either make it possible for others to replicate the model with the same dataset, or provide access to the model. In general. releasing code and data is often one good way to accomplish this, but reproducibility can also be provided via detailed instructions for how to replicate the results, access to a hosted model (e.g., in the case of a large language model), releasing of a model checkpoint, or other means that are appropriate to the research performed.
        \item While NeurIPS does not require releasing code, the conference does require all submissions to provide some reasonable avenue for reproducibility, which may depend on the nature of the contribution. For example
        \begin{enumerate}
            \item If the contribution is primarily a new algorithm, the paper should make it clear how to reproduce that algorithm.
            \item If the contribution is primarily a new model architecture, the paper should describe the architecture clearly and fully.
            \item If the contribution is a new model (e.g., a large language model), then there should either be a way to access this model for reproducing the results or a way to reproduce the model (e.g., with an open-source dataset or instructions for how to construct the dataset).
            \item We recognize that reproducibility may be tricky in some cases, in which case authors are welcome to describe the particular way they provide for reproducibility. In the case of closed-source models, it may be that access to the model is limited in some way (e.g., to registered users), but it should be possible for other researchers to have some path to reproducing or verifying the results.
        \end{enumerate}
    \end{itemize}

\item {\bf Open access to data and code}
    \item[] Question: Does the paper provide open access to the data and code, with sufficient instructions to faithfully reproduce the main experimental results, as described in supplemental material?
    \item[] Answer: \answerYes{} % Replace by \answerYes{}, \answerNo{}, or \answerNA{}.
    \item[] Justification: Given that the Dataset \& Benchmark Track adopts a single-blind review process, all related code has been publicly released on GitHub. The accompanying repository includes detailed documentation and usage guidelines to help users quickly get started with our framework.
    \item[] Guidelines:
    \begin{itemize}
        \item The answer NA means that paper does not include experiments requiring code.
        \item Please see the NeurIPS code and data submission guidelines (\url{https://nips.cc/public/guides/CodeSubmissionPolicy}) for more details.
        \item While we encourage the release of code and data, we understand that this might not be possible, so “No” is an acceptable answer. Papers cannot be rejected simply for not including code, unless this is central to the contribution (e.g., for a new open-source benchmark).
        \item The instructions should contain the exact command and environment needed to run to reproduce the results. See the NeurIPS code and data submission guidelines (\url{https://nips.cc/public/guides/CodeSubmissionPolicy}) for more details.
        \item The authors should provide instructions on data access and preparation, including how to access the raw data, preprocessed data, intermediate data, and generated data, etc.
        \item The authors should provide scripts to reproduce all experimental results for the new proposed method and baselines. If only a subset of experiments are reproducible, they should state which ones are omitted from the script and why.
        \item At submission time, to preserve anonymity, the authors should release anonymized versions (if applicable).
        \item Providing as much information as possible in supplemental material (appended to the paper) is recommended, but including URLs to data and code is permitted.
    \end{itemize}

\item {\bf Experimental setting/details}
    \item[] Question: Does the paper specify all the training and test details (e.g., data splits, hyperparameters, how they were chosen, type of optimizer, etc.) necessary to understand the results?
    \item[] Answer: \answerYes{} % Replace by \answerYes{}, \answerNo{}, or \answerNA{}.
    \item[] Justification: Key parameters such as the number of epochs, batch size, and training steps are explicitly reported in the paper. Additional training details can be found in the configuration files provided in the code repository.
    \item[] Guidelines:
    \begin{itemize}
        \item The answer NA means that the paper does not include experiments.
        \item The experimental setting should be presented in the core of the paper to a level of detail that is necessary to appreciate the results and make sense of them.
        \item The full details can be provided either with the code, in appendix, or as supplemental material.
    \end{itemize}

\item {\bf Experiment statistical significance}
    \item[] Question: Does the paper report error bars suitably and correctly defined or other appropriate information about the statistical significance of the experiments?
    \item[] Answer: \answerNo{} % Replace by \answerYes{}, \answerNo{}, or \answerNA{}.
    \item[] Justification: The paper includes extensive experimental results. However, as a benchmark-focused study, it does not require statistical significance testing in the traditional sense. Instead, we report absolute errors relative to the original results and provide detailed analysis and discussion of these discrepancies.

    \item[] Guidelines:
    \begin{itemize}
        \item The answer NA means that the paper does not include experiments.
        \item The authors should answer "Yes" if the results are accompanied by error bars, confidence intervals, or statistical significance tests, at least for the experiments that support the main claims of the paper.
        \item The factors of variability that the error bars are capturing should be clearly stated (for example, train/test split, initialization, random drawing of some parameter, or overall run with given experimental conditions).
        \item The method for calculating the error bars should be explained (closed form formula, call to a library function, bootstrap, etc.)
        \item The assumptions made should be given (e.g., Normally distributed errors).
        \item It should be clear whether the error bar is the standard deviation or the standard error of the mean.
        \item It is OK to report 1-sigma error bars, but one should state it. The authors should preferably report a 2-sigma error bar than state that they have a 96\% CI, if the hypothesis of Normality of errors is not verified.
        \item For asymmetric distributions, the authors should be careful not to show in tables or figures symmetric error bars that would yield results that are out of range (e.g. negative error rates).
        \item If error bars are reported in tables or plots, The authors should explain in the text how they were calculated and reference the corresponding figures or tables in the text.
    \end{itemize}

\item {\bf Experiments compute resources}
    \item[] Question: For each experiment, does the paper provide sufficient information on the computer resources (type of compute workers, memory, time of execution) needed to reproduce the experiments?
    \item[] Answer: \answerYes{} % Replace by \answerYes{}, \answerNo{}, or \answerNA{}.
    \item[] Justification: The paper includes a comprehensive list of the computational resources and specific parameters used to conduct our experiments.

    \item[] Guidelines:
    \begin{itemize}
        \item The answer NA means that the paper does not include experiments.
        \item The paper should indicate the type of compute workers CPU or GPU, internal cluster, or cloud provider, including relevant memory and storage.
        \item The paper should provide the amount of compute required for each of the individual experimental runs as well as estimate the total compute. 
        \item The paper should disclose whether the full research project required more compute than the experiments reported in the paper (e.g., preliminary or failed experiments that didn't make it into the paper). 
    \end{itemize}
    
\item {\bf Code of ethics}
    \item[] Question: Does the research conducted in the paper conform, in every respect, with the NeurIPS Code of Ethics \url{https://neurips.cc/public/EthicsGuidelines}?
    \item[] Answer: \answerYes{}{} % Replace by \answerYes{}, \answerNo{}, or \answerNA{}.
    \item[] Justification: he research strictly follows the NeurIPS Code of Ethics. It involves only benchmark evaluations on publicly available datasets (e.g., CIFAR-10/100) without using any personally identifiable information or human-related data. No ethical concerns are identified in this work.
    \item[] Guidelines:
    \begin{itemize}
        \item The answer NA means that the authors have not reviewed the NeurIPS Code of Ethics.
        \item If the authors answer No, they should explain the special circumstances that require a deviation from the Code of Ethics.
        \item The authors should make sure to preserve anonymity (e.g., if there is a special consideration due to laws or regulations in their jurisdiction).
    \end{itemize}

\item {\bf Broader impacts}
    \item[] Question: Does the paper discuss both potential positive societal impacts and negative societal impacts of the work performed?
    \item[] Answer: \answerNo{} % Replace by \answerYes{}, \answerNo{}, or \answerNA{}.
    \item[] Justification: This study aims to explore the performance and limitations of Spiking Transformers, with the goal of providing insights and recommendations for the future development of Spiking Transformers and related models. The work does not involve any societal or ethical impact.

    \item[] Guidelines:
    \begin{itemize}
        \item The answer NA means that there is no societal impact of the work performed.
        \item If the authors answer NA or No, they should explain why their work has no societal impact or why the paper does not address societal impact.
        \item Examples of negative societal impacts include potential malicious or unintended uses (e.g., disinformation, generating fake profiles, surveillance), fairness considerations (e.g., deployment of technologies that could make decisions that unfairly impact specific groups), privacy considerations, and security considerations.
        \item The conference expects that many papers will be foundational research and not tied to particular applications, let alone deployments. However, if there is a direct path to any negative applications, the authors should point it out. For example, it is legitimate to point out that an improvement in the quality of generative models could be used to generate deepfakes for disinformation. On the other hand, it is not needed to point out that a generic algorithm for optimizing neural networks could enable people to train models that generate Deepfakes faster.
        \item The authors should consider possible harms that could arise when the technology is being used as intended and functioning correctly, harms that could arise when the technology is being used as intended but gives incorrect results, and harms following from (intentional or unintentional) misuse of the technology.
        \item If there are negative societal impacts, the authors could also discuss possible mitigation strategies (e.g., gated release of models, providing defenses in addition to attacks, mechanisms for monitoring misuse, mechanisms to monitor how a system learns from feedback over time, improving the efficiency and accessibility of ML).
    \end{itemize}
    
\item {\bf Safeguards}
    \item[] Question: Does the paper describe safeguards that have been put in place for responsible release of data or models that have a high risk for misuse (e.g., pretrained language models, image generators, or scraped datasets)?
    \item[] Answer: \answerNo{} % Replace by \answerYes{}, \answerNo{}, or \answerNA{}.
    \item[] Justification: The work does not involve the release of any models or datasets that carry a high risk of misuse. It purely benchmarks existing open-source models on public datasets.
    \item[] Guidelines:
    \begin{itemize}
        \item The answer NA means that the paper poses no such risks.
        \item Released models that have a high risk for misuse or dual-use should be released with necessary safeguards to allow for controlled use of the model, for example by requiring that users adhere to usage guidelines or restrictions to access the model or implementing safety filters. 
        \item Datasets that have been scraped from the Internet could pose safety risks. The authors should describe how they avoided releasing unsafe images.
        \item We recognize that providing effective safeguards is challenging, and many papers do not require this, but we encourage authors to take this into account and make a best faith effort.
    \end{itemize}

\item {\bf Licenses for existing assets}
    \item[] Question: Are the creators or original owners of assets (e.g., code, data, models), used in the paper, properly credited and are the license and terms of use explicitly mentioned and properly respected?
    \item[] Answer: \answerYes{} % Replace by \answerYes{}, \answerNo{}, or \answerNA{}.
    \item[] Justification: We properly cite the original papers for all datasets (e.g., CIFAR-10/100) and models used in this study, and respect their corresponding licenses and terms of use.
    \item[] Guidelines:
    \begin{itemize}
        \item The answer NA means that the paper does not use existing assets.
        \item The authors should cite the original paper that produced the code package or dataset.
        \item The authors should state which version of the asset is used and, if possible, include a URL.
        \item The name of the license (e.g., CC-BY 4.0) should be included for each asset.
        \item For scraped data from a particular source (e.g., website), the copyright and terms of service of that source should be provided.
        \item If assets are released, the license, copyright information, and terms of use in the package should be provided. For popular datasets, \url{paperswithcode.com/datasets} has curated licenses for some datasets. Their licensing guide can help determine the license of a dataset.
        \item For existing datasets that are re-packaged, both the original license and the license of the derived asset (if it has changed) should be provided.
        \item If this information is not available online, the authors are encouraged to reach out to the asset's creators.
    \end{itemize}

\item {\bf New assets}
    \item[] Question: Are new assets introduced in the paper well documented and is the documentation provided alongside the assets?
    \item[] Answer: \answerNA{} % Replace by \answerYes{}, \answerNo{}, or \answerNA{}.
    \item[] Justification: The paper does not introduce or release any new datasets, models, or code assets.
    \item[] Guidelines:
    \begin{itemize}
        \item The answer NA means that the paper does not release new assets.
        \item Researchers should communicate the details of the dataset/code/model as part of their submissions via structured templates. This includes details about training, license, limitations, etc. 
        \item The paper should discuss whether and how consent was obtained from people whose asset is used.
        \item At submission time, remember to anonymize your assets (if applicable). You can either create an anonymized URL or include an anonymized zip file.
    \end{itemize}

\item {\bf Crowdsourcing and research with human subjects}
    \item[] Question: For crowdsourcing experiments and research with human subjects, does the paper include the full text of instructions given to participants and screenshots, if applicable, as well as details about compensation (if any)? 
    \item[] Answer: \answerNA{} % Replace by \answerYes{}, \answerNo{}, or \answerNA{}.
    \item[] Justification: The paper does not involve any crowdsourcing experiments or research with human participants.
    \item[] Guidelines:
    \begin{itemize}
        \item The answer NA means that the paper does not involve crowdsourcing nor research with human subjects.
        \item Including this information in the supplemental material is fine, but if the main contribution of the paper involves human subjects, then as much detail as possible should be included in the main paper. 
        \item According to the NeurIPS Code of Ethics, workers involved in data collection, curation, or other labor should be paid at least the minimum wage in the country of the data collector. 
    \end{itemize}

\item {\bf Institutional review board (IRB) approvals or equivalent for research with human subjects}
    \item[] Question: Does the paper describe potential risks incurred by study participants, whether such risks were disclosed to the subjects, and whether Institutional Review Board (IRB) approvals (or an equivalent approval/review based on the requirements of your country or institution) were obtained?
    \item[] Answer: \answerNA{}{} % Replace by \answerYes{}, \answerNo{}, or \answerNA{}.
    \item[] Justification: The paper does not involve any research with human participants and thus does not require IRB approval.
    \item[] Guidelines:
    \begin{itemize}
        \item The answer NA means that the paper does not involve crowdsourcing nor research with human subjects.
        \item Depending on the country in which research is conducted, IRB approval (or equivalent) may be required for any human subjects research. If you obtained IRB approval, you should clearly state this in the paper. 
        \item We recognize that the procedures for this may vary significantly between institutions and locations, and we expect authors to adhere to the NeurIPS Code of Ethics and the guidelines for their institution. 
        \item For initial submissions, do not include any information that would break anonymity (if applicable), such as the institution conducting the review.
    \end{itemize}

\item {\bf Declaration of LLM usage}
    \item[] Question: Does the paper describe the usage of LLMs if it is an important, original, or non-standard component of the core methods in this research? Note that if the LLM is used only for writing, editing, or formatting purposes and does not impact the core methodology, scientific rigorousness, or originality of the research, declaration is not required.
    %this research? 
    \item[] Answer: \answerNA{} % Replace by \answerYes{}, \answerNo{}, or \answerNA{}.
    \item[] Justification: The paper does not involve the use of LLMs as any important, original, or non-standard component in the development of the core methodology.
    \item[] Guidelines:
    \begin{itemize}
        \item The answer NA means that the core method development in this research does not involve LLMs as any important, original, or non-standard components.
        \item Please refer to our LLM policy (\url{https://neurips.cc/Conferences/2025/LLM}) for what should or should not be described.
    \end{itemize}

\end{enumerate}

\newpage
\appendix
\section{Spiking Transformer Achitectures}

\subsection{Spiking Encoding}
\label{ap:encoding}
\begin{figure}[h]
\centering
\includegraphics[width=1\textwidth]{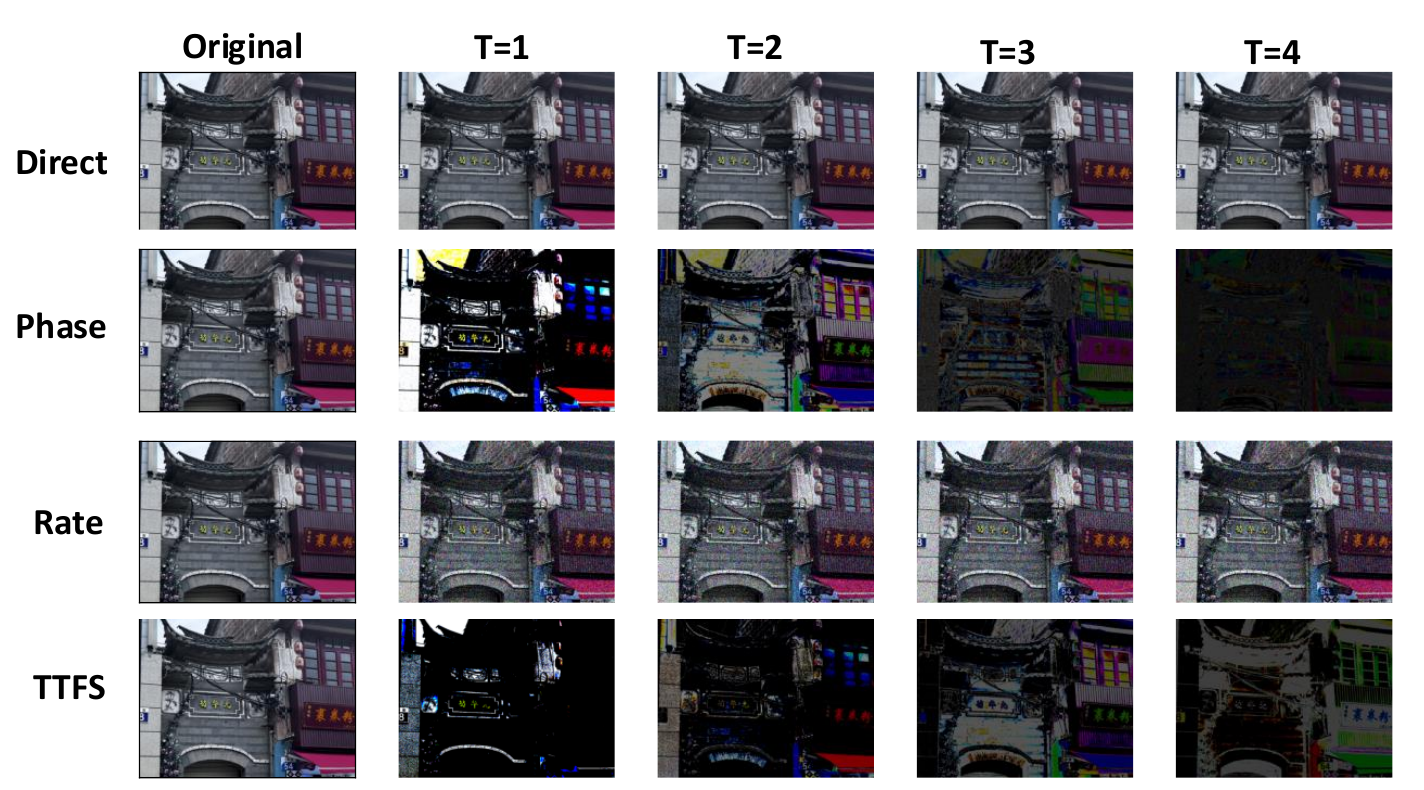}
\caption{Visualization of different encoding methods.}
\label{fig:encoding}
\end{figure}

Fig.\ref{fig:encoding} illustrates the four spike-based input encoding schemes used in our study: Direct, Phase\cite{kim2018deep}, Rate~\cite{adrian1926impulses}, and Time-to-First-Spike (TTFS)~\cite{park2020t2fsnn}. For each static frame, we visualize its transformation over four discrete simulation steps (T=1…4), showing how pixel intensities are mapped into temporally distributed spikes through different strategies. Direct encoding preserves raw intensity at every step, Phase encoding modulates spike timing periodically, Rate encoding converts intensity to firing frequency, and TTFS uses the latency of the first spike to encode information. These complementary methods introduce diverse temporal input dynamics for our Spiking Transformer benchmark, enabling fair model evaluation under varied temporal signal structures. The specific formulations for these encodings can be found in Eq.~\ref{eq:encoding_direct}-~\ref{eq:encoding_ttfs}.

\paragraph{Direct Encoding}
\begin{equation}
\label{eq:encoding_direct}
S_t(\mathbf{p}) \;=\; x(\mathbf{p}), 
\qquad t = 1,\dots,T ,
\end{equation}
where $x(\mathbf{p})\!\in\![0,1]$ denotes the normalized pixel (or feature) intensity at spatial coordinate $\mathbf{p}$.  
The same constant input current is injected at every time step, i.e.\ the spike train is temporally uniform.

\paragraph{Phase Encoding}
\begin{equation}
S_t(\mathbf{p}) \;=\;
\begin{cases}
2^{-\,(b+1)}, & \text{if } v_{\,7-b}(\mathbf{p}) = 1,\; b \equiv (t-1) \pmod{8},\\[4pt]
0, & \text{otherwise},
\end{cases}
\qquad 
v(\mathbf{p}) = \bigl\lfloor 256\,x(\mathbf{p}) \bigr\rfloor ,
\end{equation}
where $v_{k}$ is the $k$-th bit of the 8-bit integer $v(\mathbf{p})$ (most significant bit $k=7$).  
The encoder cycles through the eight bit-planes, assigning a weight that halves with each less-significant bit.

\paragraph{Rate Encoding}
\begin{equation}
S_t(\mathbf{p}) \;\sim\; \operatorname{Bernoulli}\!\bigl(x(\mathbf{p})\bigr),
\qquad 
\mathbb{E}\!\bigl[S_t(\mathbf{p})\bigr] \;=\; x(\mathbf{p}),
\qquad t = 1,\dots,T .
\end{equation}
A spike is emitted at time $t$ with probability proportional to the input magnitude, so that the average firing rate reflects $x(\mathbf{p})$.

\paragraph{Time-to-First-Spike (TTFS) Encoding}
\begin{align}
\label{eq:encoding_ttfs}
t^{\star}(\mathbf{p}) &= 1 + \Bigl\lfloor \bigl(1 - x(\mathbf{p})\bigr)\,T \Bigr\rfloor, \\[4pt]
S_t(\mathbf{p}) &= 
\begin{cases}
\dfrac{1}{t^{\star}(\mathbf{p})}, & t = t^{\star}(\mathbf{p}),\\[8pt]
0, & \text{otherwise}.
\end{cases}
\end{align}
Each neuron fires exactly once; higher input values trigger earlier spikes.  
We scale the spike amplitude by $1/t^{\star}$ to preserve energy across different latencies, but a binary value 1 can be used instead if desired.

\subsection{Spiking Neuron}
\label{ap:neuron}
\begin{figure}[h]
\centering
\includegraphics[width=1\textwidth]{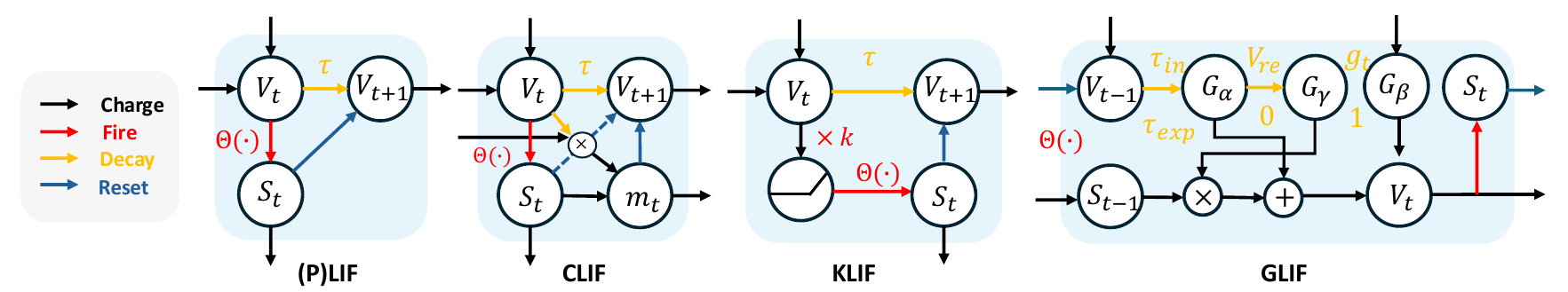}
\caption{Visualization of different spiking neurons used in this work.}
\label{fig:neuron}
\end{figure}

In this appendix, we describe the five discrete-time spiking neuron models integrated into our benchmark Spiking Transformer and summarize their defining characteristics. The vanilla LIF model implements the classical leak–fire–reset cycle, accumulating synaptic input and decaying with a fixed time constant $\tau$  before emitting a spike via a hard threshold \cite{hunsberger2015spiking}. PLIF extends this formulation by introducing a learnable membrane time constant for each neuron, allowing decay dynamics to adapt during training \cite{fang2021incorporating}. Building on PLIF, CLIF incorporates a complementary trace variable that smooths the surrogate-gradient around threshold crossings, thereby improving gradient flow in deep SNNs \cite{huang2024clif}. GLIF further enriches membrane dynamics with multiple gated internal states, capturing adaptation and refractory processes to more closely mimic biological neurons \cite{yao2022glif}. Finally, KLIF replaces the standard exponential leak with a learnable kernel constant, optimizing decay behavior for both biological realism and computational efficiency \cite{jiang2023klif}. Section 4 reports the classification accuracy of each variant, enabling a comparative analysis of how these neuron-level enhancements influence Spiking Transformer performance and hardware requirements. The specific structure of neuron can be viewed in Fig.~\ref{fig:neuron}.

\subsection{Model Basic Achitecture}
\label{ap: ST Arch}
With only a few specialised variants as exceptions, the architecture of the Spiking Transformer can be succinctly formalised by the following equations:
\begin{equation}
    \begin{gathered}
        X = \text{SPS}(\mathrm{Input}),\ \text{PE} = \text{SN}(\text{BN}(\text{Conv2d}(X))), \\
        X_0 = x + \text{PE}, \\
        X'_l = \text{Spiking Attn}(X_{l-1}) + X_{l-1}, \ X_l = \text{MLP}(X'_l) + X'_l, \\
        Y = \text{Heads}(\text{AP}(X_L))
    \end{gathered}
    \label{eq:ST arch}
\end{equation}
Eq.~\ref{eq:ST arch} factorises the pipeline into four stages.  
\textbf{(1) SPS.} A four-stage Sequential Patch Splitting module—each stage stacks \texttt{Conv}–\texttt{BN}–\texttt{Pool}–Spiking-Neuron—upsamples the raw input into token feature maps \(X\).  
\textbf{(2) Positional Encoding.} A shallow \texttt{Conv} followed by \texttt{BN} and a spiking activation produces \(\text{PE}\), which is added to \(X\) to obtain the embedded sequence \(X_0\).  
\textbf{(3) Transformer Block.} \(L\) residual blocks alternate \textit{Spiking Self-Attention} and \textit{MLP} layers, yielding hidden states \(\{X_l\}_{l=1}^{L}\).  
\textbf{(4) Head.} Global average pooling \(\text{AP}(\cdot)\) followed by a task-specific head maps the final representation to the output \(Y\).

Together, SPS and positional encoding realise the input-to-embedding conversion, while the stacked spiking blocks capture spatiotemporal dependencies with neuromorphic efficiency.

\subsection{Spiking Attention}
\label{ap:Attn}
We select Spikformer, Spike-driven Transformer, and Spikformer+SEMM as three representative models. Detailed descriptions of the Attention mechanisms used in the latter two are provided below.
\paragraph{SDSA}
In Eq.~\ref{eq:SDSA}, \(Q,K,V\in\mathbb{R}^{B\times N\times C}\) denote the query, key, and value tensors for a batch of size \(B\) with \(N\) tokens and \(C\) channels. The operator \(\otimes\) is an element-wise outer product between \(Q\) and \(K\); \(\operatorname{SUM}_{c}(\cdot)\) sums this product across the channel dimension \(C\). \(\operatorname{SN}(\cdot)\) is a spiking-neuron activation that returns a binary spike map. The final SDSA output is obtained by the element-wise product of this map with the value tensor \(V\).

\begin{gather}
\label{eq:SDSA}  
    SDSA(Q,K,V)= SN (SUM_c (Q \otimes K)) \otimes V 
\end{gather}

\paragraph{Spikformer+SEMM}
In Eq.~\ref{eq:SEMM}, \(m\) is the number of experts. For each expert \(i\in\{1,\dots,m\}\), \(Q_{m}\) is its private query tensor and \(A_{m}=\operatorname{SSA}_{m}(Q_{m},K,V)\) is the corresponding sub-attention result. The input feature tensor is \(X\), and \(W_{R}^{\top}\) is the router’s weight matrix. \(\operatorname{BN}(\cdot)\) applies batch normalisation, while \(\operatorname{SN}(\cdot)\) converts the routed signal into a set of spiking coefficients \(\{r_{1},r_{2},\dots,r_{m}\}\). These coefficients weight the expert outputs to form the final mixture: \(\operatorname{SSA{+}SEMM}=\sum_{i=1}^{m} r_{i} A_{i}\).
\begin{equation}
\label{eq:SEMM}  
\begin{gathered}
        A_m = SSA_m(Q_m,K,V),\quad  Router = SN\left(BN(W_R^TX)\right) = \{r_1, r_2, \ldots, r_m\} \\
        SSA+SEMM = \sum_{i=1}^{m} \mathbf{r}_i * \mathbf{A}_i,
\end{gathered}
\end{equation}

\newpage
\section{Spiking Transformer Experiments}
\subsection{Selected Spiking Transformer Performance}

We collect the performance of mainstream Spiking Transformers across a variety of static and dynamic datasets. We transcribe the original experimental results into Tab.~\ref{tab: all models} for direct comparison. For key reproduced models, we explicitly highlight their results in the tables, with detailed experimental setups and discussions available in Sec.~\ref{sec:expirements} and Sec.~\ref{sec:analysis}.

\begin{table}[!h]
    \caption{Selected Spiking Transformers \textit{A2S}: ANN-SNN Conversion Model; \textit{Transfer}: Transfer Learning Model; \textit{T}: Time Step.}
    \label{tab: all models}
    \centering
    \renewcommand{\arraystretch}{1.2} 
    \resizebox{\textwidth}{!}{
    \begin{tabular}{c c c c c c}
      \toprule
        \diagbox[width=2.5cm, linewidth=0.0pt]{\textbf{Model}}{\textbf{Datasset}} &    \textbf{CIFAR10} & \textbf{CIFAR100} & \textbf{ImageNet-1K} & \textbf{CIFAR10-DVS} & \textbf{N-Cal101} \\
    
    \midrule
      \textbf{Spikformer}~\cite{zhou2022spikformer} & 95.41 & 78.21 & 74.81 & 78.9 & - \\
      Spikformer v2~\cite{zhou2024spikformer} & - & - & 80.38 \textit{(8-512)} & - & - \\
      \textbf{QKFormer}~\cite{zhou2024qkformer} & 96.18 & 81.15 & 85.65 \textit{(10-768)} & 84.0\textit{(T=16)}& - \\
      \textbf{Spikingformer}~\cite{zhou2023spikingformer} & 95.81 & 79.21 & 75.85 & 79.9 & - \\
      \textbf{SGLFormer}~\cite{zhang2024sglformer} & 96.76 & 82.26 & 83.73 & 82.9 & - \\
      \textbf{Spiking Wavelet Transformer}~\cite{fang2024spiking} & 96.1 & 79.3 & 75.34 \textit{(8-512)} & 82.9 & 88.45 \\
      \textbf{Spike-driven Transformer}~\cite{yao2023spike} & 95.6 & 78.4 (2-512) & 77.07 & 80.0 \textit{(T=16)} & - \\
      Meta-SpikeFormer(SDT v2)~\cite{yao2024spike} & - & - & 80.00 & - & - \\
      E-SpikeFormer(SDT v3)~\cite{yao2025scaling} & - & - & 86.20 \textit{(T=8)} & - & - \\
      MST~\cite{wang2023masked} & 97.27 \textit{(A2S) }& 86.91 \textit{(A2S)}  & 78.51 \textit{(A2S)} & 88.12 \textit{(A2S)} & 91.38\textit{ (A2S)} \\
      QSD~\cite{qiu2025quantized} & 98.4 \textit{(Transfer)} & 87.6 \textit{(Transfer)} & 80.3 & 89.8 \textit{(Transfer)} & - \\
      Spiking Transformer~\cite{guo2025spiking} & 96.32 & 79.69 & 78.66 \textit{(10-512)}& - & - \\
      SNN-ViT~\cite{wang2025spiking} & 96.1 & 80.1 & 80.23 & 82.3 & - \\
      STSSA~\cite{zhou2025spiking} & - & - & - & 83.8 & 81.65 \\
      \textbf{Spikformer + SEMM}~\cite{zhou2024spiking} & 95.78 & 79.04 & 75.93 \textit{(8-512)} & 82.32 & - \\
      \textbf{SpikingResformer}~\cite{shi2024spikingresformer} & 97.40 \textit{(Transfer)} & 85.98 \textit{ (Transfer)} & 79.40 & 84.8\textit{ (Transfer)} & - \\
      TIM~\cite{shen2024tim} & - & - & - & 81.6 & 79.00 \\
      \bottomrule
    \end{tabular}
    }
  \end{table}
  
\subsection{Visualization on ImageNet-1k} \label{sec:app_imagenet}
\begin{figure}[h]
\centering
\includegraphics[width=0.65\textwidth]{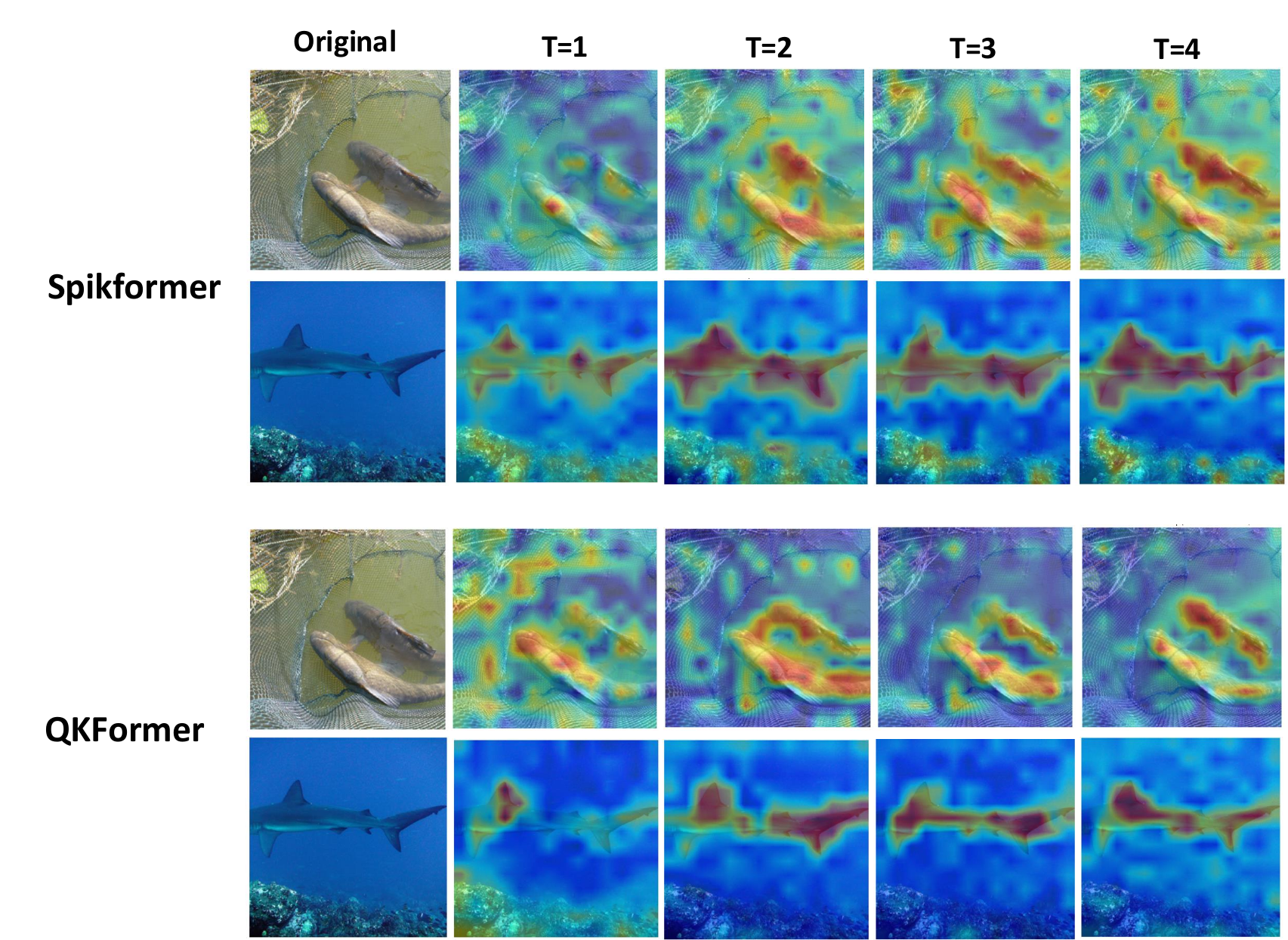}
\caption{Visualization the importance weight using GradCam++ on QKFormer and Spikformer ImageNet-1k}
\label{fig:imagenet_gradcpp}
\end{figure}

To further interpret the temporal dynamics of Spiking Transformers, we employ Grad-CAM++~\cite{chattopadhay2018grad} to visualize the attention maps across four simulation steps (T=1 to T=4) for both Spikformer~\cite{zhou2024spikformer} and QKFormer~\cite{zhou2024qkformer} on ImageNet-1k samples (Fig.~\ref{fig:imagenet_gradcpp}). These visualizations offer insight into how each model accumulates temporal evidence and localizes discriminative features over time.

QKFormer demonstrates consistent and focused attention on the object regions across all time steps, especially for challenging examples such as the shark in a low-contrast underwater scene. This indicates a stable spatial grounding and effective temporal integration, likely attributable to its hierarchical pyramid architecture that supports multiscale representation.

In contrast, Spikformer exhibits broader and more diffuse activation patterns in early time steps, gradually converging toward the object. However, its attention maps remain noisier and less confined, particularly in scenes with complex backgrounds. This suggests that while Spikformer may respond quickly to salient regions, its spatial precision is relatively limited compared to QKFormer.

Overall, these results underscore the importance of temporal consistency and multiscale design in spiking vision transformers. QKFormer’s clear and persistent localization highlights the benefit of incorporating hierarchical cues, aligning well with its superior top-1 performance.

\subsection{Result on COCO}
\label{ap:det}
\begin{figure}[h]
\centering
\includegraphics[width=0.9\textwidth]{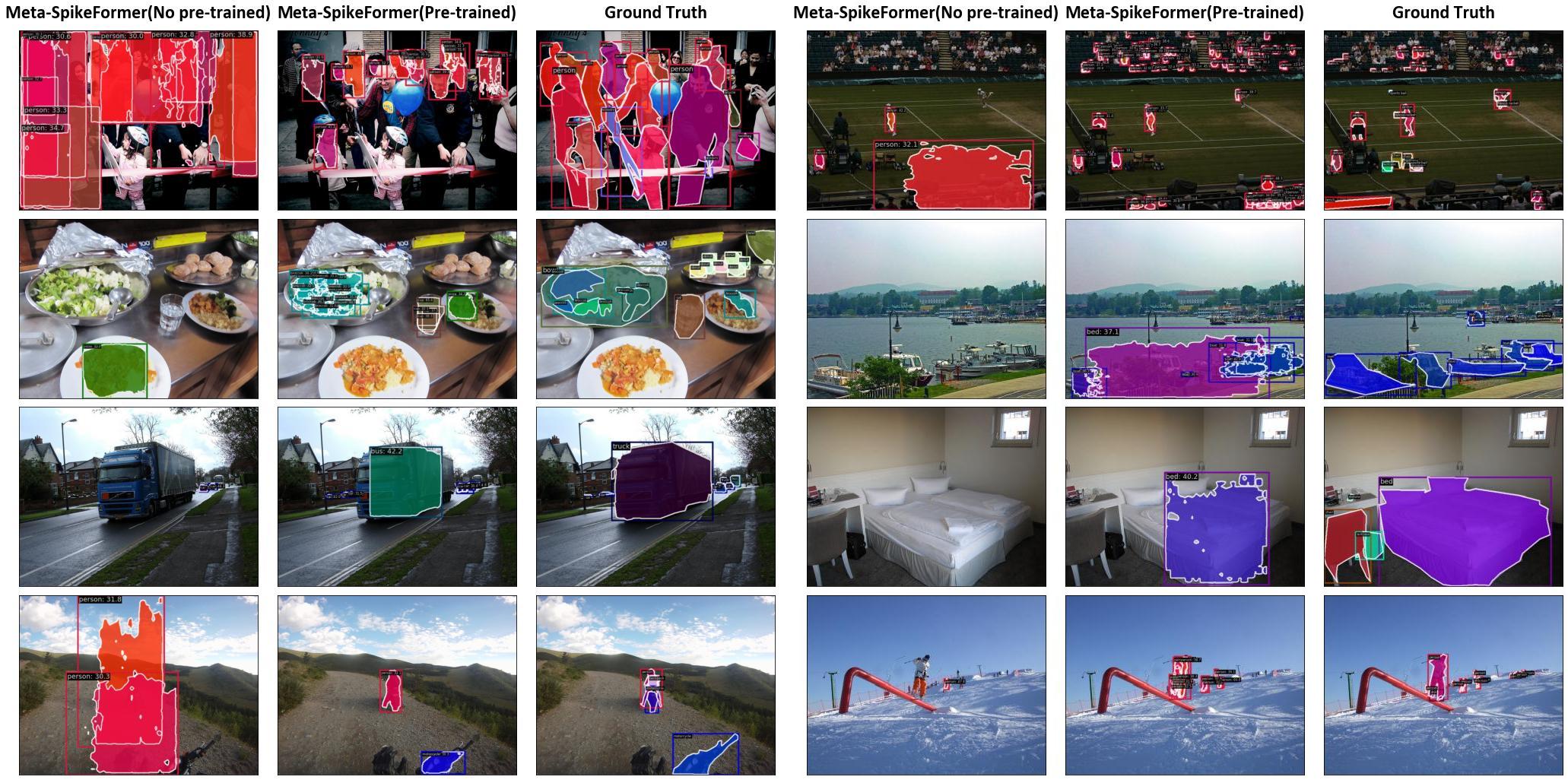}
\caption{Detection predictions on COCO for SDTv2.}
\label{fig:det_predict_large}
\end{figure}
To assess the impact of pretraining on detection performance, we adopt SDTv2~\cite{yao2024spike} as the backbone for object detection, integrating it into the Mask R-CNN framework. SDTv2 replaces the standard CNN backbone with a custom Spiking Vision Transformer, featuring embedding dimensions of [128, 256, 512, 640], 8 heads, and 8 layers per stage. Its spike-driven self-attention (SDSA) enables efficient feature extraction with reduced computational cost.

Multi-scale features extracted by SDTv2 are fused via a SpikeFPN neck. The RPN and ROI heads are adapted to the spiking domain using SpikeRPNHead and SpikeStandardRoIHead, preserving spike-driven computation throughout the pipeline.

Training follows a standard augmentation regime (random flipping, resizing) and a linear warm-up/decay schedule. We use AdamW with a learning rate of 2.5e-5, betas (0.9, 0.999), and weight decay of 0.05.

On COCO, SDTv2 achieves competitive performance with significantly lower power consumption. As shown in Fig.~\ref{fig:det_predict_large}, visual comparisons further illustrate the benefit of pretraining, highlighting the suitability of SDTv2 for efficient, neuromorphic detection systems.

\subsection{Framework Robustness}
\label{sec:robustness}
To demonstrate the robustness of the framework, we subsequently tested the main models on the primary datasets. The results in Tab.~\ref{tab:robutness} show that the models reproduced based on our framework maintained both standard deviation and confidence intervals within a reasonable range across multiple datasets, including both static and neuromorphic ones.
\begin{table*}[h!]
\centering
\caption{Classification accuracy and confidence intervals on CIFAR10, CIFAR100, and CIFAR10-DVS datasets.}
 \resizebox{0.95\linewidth}{!}{
\begin{tabular}{lcccc}
\hline
\textbf{Model} & \textbf{Dataset} & \textbf{Acc@1 (\%)} & \textbf{Std (\%)} & \textbf{t-95\% Confidence Interval} \\
\hline
Spikformer & CIFAR10 & 95.16 & 0.09 & [95.04\%, 95.27\%] \\
SDT & CIFAR10 & 95.77 & 0.04 & [95.72\%, 95.82\%] \\
QKFormer & CIFAR10 & 96.21 & 0.03 & [96.18\%, 96.25\%] \\
Spikformer+SEMM & CIFAR10 & 95.55 & 0.33 & [95.14\%, 95.96\%] \\
Spikingformer & CIFAR10 & 95.47 & 0.11 & [95.32\%, 95.61\%] \\
\hline
Spikformer & CIFAR100 & 77.76 & 0.31 & [77.38\%, 78.14\%] \\
SDT & CIFAR100 & 78.37 & 0.14 & [78.19\%, 78.54\%] \\
QKFormer & CIFAR100 & 79.95 & 0.18 & [79.72\%, 80.17\%] \\
Spikformer+SEMM & CIFAR100 & 78.41 & 0.52 & [77.76\%, 79.06\%] \\
Spikingformer & CIFAR100 & 79.33 & 0.21 & [79.06\%, 79.59\%] \\
\hline
Spikformer & CIFAR10-DVS & 83.40 & 1.76 & [81.12\%, 85.49\%] \\
SDT & CIFAR10-DVS & 80.33 & 0.60 & [79.58\%, 81.08\%] \\
QKFormer & CIFAR10-DVS & 79.77 & 0.58 & [79.05\%, 80.49\%] \\
\hline
\end{tabular}}
\label{tab:robutness}
\end{table*}

\subsection{Complicated Datasets}
\label{sec:Complicated Datasets}
In addition to traditional tasks such as classification, segmentation, and detection, the capabilities of the Spiking Transformer have also been generalized to other datasets. These datasets include 3D point cloud classification tasks such as ModelNet10/40, as well as event-based video detection using DVS data. However, processing these datasets requires model-specific optimization, meaning that some basic baselines cannot be quickly adapted to support these tasks. Nevertheless, our framework integrates them to facilitate development and research for users.
To demonstrate the successful integration of these tasks within our framework, we obtained preliminary results using the corresponding models on their respective tasks:

\begin{table}[h!]
\centering
 \resizebox{1\linewidth}{!}{
\begin{tabular}{lcccc}
\toprule
\textbf{model} & \textbf{mAP@[0.50:0.95]} & \textbf{mAP@0.5} & \textbf{AR@[0.50:0.95] (all)} & \textbf{AR@[0.50:0.95] (large)} \\
\hline
\textbf{SDT}~\cite{yao2023spike} & 0.000 & 0.001 & 0.008 & 0.028 \\
\textbf{SODFormer}~\cite{li2023sodformer} & 0.000 & 0.001 & 0.023 & 0.034 \\
\bottomrule
\end{tabular}}
\caption{Spike\-Driven Transformer \& SODFormer(baseline) on PKU\_DAVIS\_SOD with 3 epochs}
\label{tab:pku}
\end{table}

\begin{table}[h!]
\centering
\begin{tabular}{lcccc}
\toprule
\textbf{Model} & \textbf{Dataset} & \textbf{Step} & \textbf{Acc@1(\%)} & \textbf{Epoch} \\
\hline
Spiking Point Transformer~\cite{wu2025spiking} & ModelNet10 & 4 & 90.5 & 200 \\
\bottomrule
\end{tabular}
\caption{Performance of Spiking Point Transformer on the ModelNet10 dataset}
\label{tab:spt_modelnet10}
\end{table}

According to Tab.~\ref{tab:pku}, SDT exhibits a significant performance gap compared to the baseline on tasks without specific adaptation. In addition, it should be noted that the results of SODFormer and Spiking Point Transformer are presented merely to demonstrate the framework’s compatibility with different datasets and tasks; their parameters are not fully aligned with those in the original papers and therefore do not reflect the actual performance of the models.

\newpage
\section{STEP Quick Start}
\label{ap:step_quickstart}
\subsection{STEP Structure Overview}
STEP is a modular benchmark framework designed for multi-task evaluation. It features a well-structured architecture while maintaining strong accessibility for users. The core structure of the STEP codebase is organized as follows:
\noindent
\begin{RoundedBox}{STEP Repo Structure}
\ttfamily
STEP/\\
+-- cls/               \qquad \# Classification submodule\\
|   +-- README.md\\
|   +-- configs/\\
|   +-- datasets/\\
|   +-- \dots\\
+-- seg/             \qquad  \# Segmentation submodule\\
|   +-- README.md\\
|   +-- configs/\\
|   +-- mmseg/\\
|   +-- \dots\\
+-- det/              \qquad \# Object detection submodule\\
|   +-- README.md \\
|   +-- configs/ \\
|   +-- mmdet/ \\
|   +-- \dots
\end{RoundedBox}

\subsection{Classification Demo}
In STEP, once components such as attention modules, neuron models, or encoding schemes are implemented, a complete model is assembled via a configuration file (.yml file per model setting), which then initiates the training pipeline. 

For the classification task, the model can be configured using a configuration file as shown below. Here, we take the example of Spikformer evaluated on the CIFAR-10 dataset:
\noindent
\begin{RoundedBox}{Spikformer CIFAR-10 Config}
\ttfamily
\# dataset\\
data\_dir: '/data/datasets/CIFAR10'\\
dataset: torch/cifar10\\
num\_classes: 10\\
img\_size: 32\\
\\
\# data augmentation\\
mean:\\
\hspace*{1em}- 0.4914\\
\hspace*{1em}- 0.4822\\
\hspace*{1em}- 0.4465\\
std:\\
\hspace*{1em}- 0.2470\\
\hspace*{1em}- 0.2435\\
\hspace*{1em}- 0.2616\\
crop\_pct: 1.0\\
mixup: 0.5\\
cutmix: 0.0\\
reprob: 0.25\\
remode: const\\
...
\end{RoundedBox}

\begin{RoundedBox}{Spikformer CIFAR-10 Config (Continuous)}
\ttfamily
\# model structure\\
model: "spikformer\_cifar"\\
step: 4\\
patch\_size: 4\\
in\_channels: 3\\
embed\_dim: 384\\
num\_heads: 12\\
mlp\_ratio: 4\\
depths: 4\\
\\
\# meta transformer layer\\
embed\_layer: 'SPS'\\
attn\_layer: 'SSA'\\
\\
\# node\\
tau: 2.0\\
threshold: 1.0\\
act\_function: SigmoidGrad\\
node\_type: LIFNode\\
alpha: 4.0\\
\\
\# train hyperparam\\
amp: True\\
batch\_size: 128\\
val\_batch\_size: 128\\
lr: 5e-4\\
min\_lr: 1e-5\\
sched: cosine\\
... \\
\# log dir\\
output: ".output/cls/Spikformer"\\
\# device\\
device: 0
\end{RoundedBox}

After assembly, the training script can be launched directly from the terminal. In our configuration, multiple scripts can be defined for each model to facilitate controlled, multi-round comparative experiments.
\begin{tcolorbox}[
  colback=myYellow,
  colframe=myYellow2,
  boxrule=0.4pt,
  arc=2pt,
  fonttitle=\bfseries\small,
  title={Example Bash Command},
  left=4pt, right=4pt, top=4pt, bottom=4pt
]
\ttfamily
    conda activate [your\_env] \\
    python train.py \-\-config configs/spikformer/cifar10.yml
\end{tcolorbox}
The above command launches the training process for a single model. For ImageNet, our models support multi-GPU training, which can be enabled by modifying the corresponding settings in the configuration file.

\subsection{Segmentation \& Detection Demo}
hese two tasks are implemented based on the MMSegmentation and MMDetection frameworks. For models already constructed in the classification module, code can be directly migrated to the corresponding task directory with minimal modification. Given the computational demands of segmentation and detection, multi-GPU training is enabled by default. The configuration structure for these tasks is largely similar to that of classification and is therefore omitted here for brevity. The corresponding task can be launched using the following command:
\begin{tcolorbox}[
  colback=myYellow,
  colframe=myYellow2,
  boxrule=0.4pt,
  arc=2pt,
  fonttitle=\bfseries\small,
  title={Example Bash Command},
  left=4pt, right=4pt, top=4pt, bottom=4pt
]
\ttfamily
    conda activate [your\_env] \\
    cd tools\\
    CUDA\_VISIBLE\_DEVICES=0,1 ./dist\_train.sh ../configs/spikformer\_8-512.py 2
\end{tcolorbox}

\subsection{Visualization}
In addition, we provide model visualization support base on GradCam++. You may load pretrained weights at any time and insert hooks at the appropriate locations to visualize internal representations or dynamic behaviors of the model:
\begin{tcolorbox}[
  colback=myYellow,
  colframe=myYellow2,
  boxrule=0.4pt,
  arc=2pt,
  fonttitle=\bfseries\small,
  title={Example Bash Command},
  left=4pt, right=4pt, top=4pt, bottom=4pt
]
\ttfamily
    conda activate [your\_env] \\
    python -m cls.vis.gradcam\_vis
\end{tcolorbox}

\newpage

\end{document}